# A Refined View of Causal Graphs and Component Sizes: SP-Closed Graph Classes and Beyond

**Christer Bäckström**  christer.backstrom@liu.se
**Peter Jonsson**  peter.jonsson@liu.se
*Department of Computer Science*
*Linköping University*
*SE-581 83 Linköping, Sweden*

## Abstract

The causal graph of a planning instance is an important tool for planning both in practice and in theory. The theoretical studies of causal graphs have largely analysed the computational complexity of planning for instances where the causal graph has a certain structure, often in combination with other parameters like the domain size of the variables. Chen and Giménez ignored even the structure and considered only the size of the weakly connected components. They proved that planning is tractable if the components are bounded by a constant and otherwise intractable. Their intractability result was, however, conditioned by an assumption from parameterised complexity theory that has no known useful relationship with the standard complexity classes. We approach the same problem from the perspective of standard complexity classes, and prove that planning is **NP**-hard for classes with unbounded components under an additional restriction we refer to as SP-closed. We then argue that most **NP**-hardness theorems for causal graphs are difficult to apply and, thus, prove a more general result; even if the component sizes grow slowly and the class is not densely populated with graphs, planning still cannot be tractable unless the polynomial hierachy collapses. Both these results still hold when restricted to the class of acyclic causal graphs. We finally give a partial characterization of the borderline between **NP**-hard and **NP**-intermediate classes, giving further insight into the problem.

## 1. Introduction

We will first briefly explain what a causal graph is and give a short survey of applications as well as theoretical results reported in the literature. Following that, we give an overview of the new results presented in this article.

### 1.1 Background

The causal graph for a planning instance is an explicit description of the variable dependencies that are implicitly defined by the operators. More precisely, it is a directed graph such that there is an arc from a variable $x$ to another variable $y$ if either $x$ appears in the precondition of an operator with an effect on $y$ or some operator has effects on both $x$ and $y$.

This standard definition of the causal graph can be traced back to Knoblock (1994) although he did not give it a name. He used the causal graph in the ALPINE algorithm, as a guidance for partitioning and ordering the variables in the process of automatically deriving state abstraction hierarchies. The actual name causal graph can be traced back to Williams and Nayak (1997). Their approach was both more general and more restricted than Knoblock's. On the one hand, they generalized the concept from binary variables to multi-valued variables, but on the other hand, they considered only acyclic causal graphs which implies that all operators are unary, i.e. every operator changes only one variable. The context of their work was the reactive planner Burton for onboard space-ship control. A causal model was compiled into a transition system that could be efficiently exploited by a reactive controller to choose appropriate operators to achieve given goals. The compilation was done in such a way that all operators were unary, and they claimed that this is often possible in real applications. The resulting acyclicity of the causal graph was then exploited by Burton, which traversed the graph bottom up in order to issue operators in an order consistent with their causal relationships.

Jonsson and Bäckström (1998b) also studied acyclic causal graphs, but referred to them as *dependency graphs*. They considered a subclass of such graphs having a particular structure and used this to implicitly define a corresponding class of planning instances, the 3S class. This class has the property that it is always possible to decide in polynomial time if there is a solution or not, but the solutions themselves may be of exponential length, thus necessarily taking exponential time to generate. Although only one single restricted case, the 3S class is probably the first example of relating structural properties of the causal graph to the computational complexity of planning. A more general and extensive such analysis was done by Domshlak and Dinitz (2001a), who analysed the complexity of planning for classes of instances corresponding to a number of different possible structures of acyclic causal graphs. However, their work was done in the context of multi-agent coordination and the term causal graph was never used.

The first two of these papers may be viewed as early examples of exploiting the causal graph in practice, while the latter papers form the starting point of the subsequent theoretical research into the relationships between planning complexity and the structure of causal graphs.

An important step forward in the usage of causal graphs was the paper by Helmert (2004) where he demonstrated that the causal graph is particularly useful in the context of multi-valued variables. Previous research on the complexity of planning with multi-valued variables had focussed on the structure of the domain-transition graphs for the variables (Jonsson & Bäckström, 1998a), rather than the causal graph. Helmert realized the power of using both the domain-transition graphs and the causal graph in heuristic planning. This was exploited in practice in his highly succesful Fast Downward planner (Helmert, 2006a). It translates PDDL planning instances with binary variables into a representation with multi-valued variables and then removes carefully chosen edges in the resulting causal graph to make it acyclic. The resulting causal graph is then used to compute a heuristic by hierarchically computing and composing plan lengths for subgraphs having one of the particular structures studied by Domshlak and Dinitz (2001a). Somewhat similarly, Katz and Domshlak (2010) identified subgraphs of the causal graph that have certain structures that make planning for them tractable. They exploited this to be able to use larger variables sets when constructing pattern databases. A further example of exploiting the causal graph to make planning more efficient is the paper on factored planning by Brafman and Domshlak (2006). They showed that the structure of the causal graph can be used as a guide for deciding if and how a planning instance can be solved more efficiently by dividing it into loosely coupled subinstances and use constraint processing. The basic idea of the causal graph to represent variable dependencies is, of course, quite general and not necessarily restricted to planning. For instance, Wehrle and Helmert (2009) transferred the causal graph concept to the context of model checking.

As previously mentioned, the two papers by Jonsson and Bäckström (1998b) and by Domshlak and Dinitz (2001a) can be viewed as the starting point for a successful line of research into studying the relationships between planning complexity and the structure of the causal graph. While the 3S class by Jonsson and Bäckström was a very limited special case, Domshlak and Dinitz studied classes of planning instances corresponding to a number of more general graph structures, like in-stars (aka. inverted forks), out-stars (aka. forks), directed path graphs (aka. directed chain graphs), polytrees and singly-connected DAGs. Further results followed, for instance, in articles by Brafman and Domshlak (2003), and Giménez and Jonsson (2008). The latter article additionally showed that although 3S instances can have exponential-length plans, it is possible to generate a macro representation of such a plan in polynomial time, a result they extended also to some other classes defined by the structure of the causal graph. Many of the complexity results in these papers use additional numerical parameters in conjunction with the graph structure. Examples of such parameters are the maximum domain size of the variables and the maximum in-degree of the graph. While increasing the number of possible cases to analyse, it does allow for a more fine-grained analysis in many cases. Consider for instance the case of directed path graphs. Domshlak and Dinitz (2001a) proved that it is tractable to decide if there is a plan for this case when the domains are binary, while Giménez and Jonsson (2009) proved that a domain size of 5 is sufficient to make the problem **NP**-hard. Similarly, Giménez and Jonsson (2012) proved tractability for planning instances with binary variables, a constant number of prevail conditions and where the causal graph is a polytree. Also the paper by Brafman and Domshlak (2006) fits into this line of theoretical research, exhibiting a planning algorithm that runs in time exponential in two parameters, the tree-width of the undirected version of the causal graph and the maximum number of times a variable must change value.

While most research has been based on the standard definition of causal graphs that was set already by Knoblock, although often in the generalisation to multi-valued variables, there are important exceptions. One potential problem with the standard defintion is that whenever two variables are both affected by the same operator, then the causal graph must necessarily contain cycles, which is the major reason why the focus has mainly been on planning with unary operators. In an attempt to circumvent this problem, Jonsson (2009) defined a more relaxed variant of the causal graph that does not always introduce cycles for non-unary operators, which can sometimes allow for a more fine-grained complexity analysis.

The previous results relate the structure of the causal graph to the complexity of satisficing planning, i.e. deciding if there is a plan. There has also been a corresponding branch of research relating the structure of the causal graph to the complexity of cost-optimal planning (cf., Katz & Domshlak, 2007, 2008, 2010; Katz & Keyder, 2012).

## 1.2 Our Contributions

All of the theoretical research above studies the complexity of planning based on the structure of the causal graph, and possibly other parameters like domain sizes. An important milestone that deviates from this line of research was an article by Chen and Giménez (2010) who did not even consider the structure of the causal graph but only a simple quantitative measure, the size of the weakly connected components. They proved that deciding if there is a plan can be done in polynomial time if and only if the size of the weakly connected components in the causal graph is bounded by a constant. In one sense, this is a very sharp and final result. However, the intractability result for unbounded components is conditional on the assumtion that $\mathbf{W}[1] \not\subseteq \mathbf{nu\text{-}FPT}$. This assumption relies on the theory of parameterised complexity theory and neither the complexity classes nor the assumption itself can be related to ordinary complexity classes in a clear way. Chen and Giménez acknowledge that the problems they prove conditionally intractable include **NP**-intermediate problems. Hence, we take their result as a take-off point for further investigation of how the component sizes reflect on the standard complexity classes. Since we know from Chen and Giménez that not all graph classes with unbounded components are **NP**-hard we must consider further restrictions in order to find **NP**-hard classes. We do so by adding a new type of closure property, SP-closure, which is incomparable to subset-closure but is a subset of minor-closure, and prove that planning is **NP**-hard for any SP-closed graph class with unbounded components. It should be noted that this result still holds for the class of all acyclic graphs, which is important considering the practical relevance of acyclicity previously mentioned.

While many graph classes that have been studied in the literature are indeed SP-closed, there also exists natural classes that lack this property. We present one way of handling such classes with the aid of non-uniform complexity theory. In this case, we are not able to show **NP**-hardness but we can show that the polynomial hierarchy collapses to its second level. This is a fairly general result that can be applied even when the component sizes grow very slowly and the graph class is not very densely populated with graphs. Also this result holds even if restricted to acyclic graphs. This result can be used to demonstrate clearly that complexity results for planning based only on the class of causal graphs does not necessarily have any connection to the complexity of a generic planning problem having the same class of causal graphs. This result also raises the question of where to find (preferably natural) **NP**-intermediate planning problems. Chen and Giménez state that **NP**-intermediate problems can be obtained by using methods similiar to the ones employed by Bodirsky and Grohe (2008). Such problems are hard to describe as natural, though. They are based on Ladner's (1975) diagonalization technique that removes a large fraction of input strings from a problem. It is apparently difficult to connect graph classes constructed by this technique with simple conditions on component growth. As an alternative, we show that graph classes where the component sizes grow polylogarithmically are **NP**-intermediate under the double assumption that $\mathbf{W}[1] \not\subseteq \mathbf{nu\text{-}FPT}$ and that the *exponential time hypothesis* (Impagliazzo & Paturi, 2001) holds. We also show that for every $k > 1$, there exists a class $\mathcal{G}_k$ of graphs such that component size is bounded by $|V(G)|^{1/k}$ for all $G \in \mathcal{G}_k$ and the corresponding planning problem is **NP**-hard. These results coarsely stake out the borderline between **NP**-hard and **NP**-intermediate classes.

A possible conclusion from this paper is that complexity analysis of planning based only on the structure of the causal graph is of limited value, and that additional parameters are needed to achieve more useful results. While this may be a fair conclusion in general, there are cases where the graph structure is sufficient. For instance, Katz, Hoffmann, and Domsh-lak (2013) have applied the result by Chen and Giménez (2010) in the context of so called *red-black planning*, a variant of delete relaxation for computing heuristics. Furthermore, even when the structure of the causal graph has to be combined with other parameters, it is still important to know the behaviour of each parameter in isolation.

The remainder of the article is structured as follows. In Section 2 we set the notation and terminology used for planning and for graphs, and in Section 3 we define causal graphs and structural planning in general. Section 4 contains a number of **NP**-hardness results for various special graph classes that we need for the main results. The first of the two main theorems of the article appears in Section 5, where we define the concept of SP-closed graph classes and prove that planning is **NP**-hard for such classes when the component size is unbounded. Section 6 discusses some of the problems with both the previous theorem and other similar results in the literature. As a way around these problems, our second main theorem shows that even without any closure requirements, planning is likely to be hard even when the components grow slowly and the graphs do not appear densely in the class. Section 7 contains some observations concerning the borderline between **NP**-intermediate and **NP**-hard planning problems. The article ends with a discussion section.

## 2. Preliminaries

This section sets the terminology and notation for planning and graphs used in this article. We write $|X|$ to denote the cardinality of a set $X$ or the length of a sequence $X$, i.e. the number of elements in $X$, and we write $||X||$ to denote the size of the representation of an object $X$.

### 2.1 Planning

Since this article has many connections with the one by Chen and Giménez (2010) we follow their notation and terminology for plannning, which is a notational variant of SAS$^+$ (Bäckström & Nebel, 1995).

An *instance* of the *planning problem* is a tuple $\Pi = (V, \mathsf{init}, \mathsf{goal}, A)$ whose components are defined as follows:

- $V$ is a finite set of variables, where each variable $v \in V$ has an associated finite domain $\mathsf{D}(v)$. Note that variables are not necessarily propositional, that is, $\mathsf{D}(v)$ may be any finite set. A *state* is a mapping $s$ defined on the variables $V$ such that $s(v) \in \mathsf{D}(v)$ for all $v \in V$. A *partial state* is a mapping $p$ defined on a subset $\mathsf{vars}(p)$ of the variables $V$ such that for all $v \in \mathsf{vars}(p)$, it holds that $p(v) \in \mathsf{D}(v)$, and $p$ is otherwise undefined.

- $\mathsf{init}$ is a state called the *initial state*.

- $\mathsf{goal}$ is a partial state.

- $A$ is a set of *operators*; each operator $a \in A$ consists of a *precondition* $\mathsf{pre}(a)$ and a *postcondition* $\mathsf{post}(a)$ which are both partial states. We often use the notation $\langle \mathsf{pre} \, ; \, \mathsf{post} \rangle$ to define an operator with precondition $\mathsf{pre}$ and postcondition $\mathsf{post}$. For instance, $\mathsf{a} = \langle x = 0, y = 1 \, ; \, z = 1 \rangle$ defines an operator $\mathsf{a}$ which is applicable in any state $s$ such that $s(x) = 0$ and $s(y) = 1$, and which has the effect of setting variable $z$ to 1.

When $s$ is a state or a partial state and $W$ is a subset of the variable set $V$, we write $s\!\restriction\! W$ to denote the partial state resulting from restricting $s$ to $W$. We say that a state $s$ is a *goal state* if $\mathsf{goal} = s\!\restriction\!\mathsf{vars}(\mathsf{goal})$.

We define a *plan* (for an instance $\Pi$) to be a sequence of operators $P = a_1, \ldots, a_n$. Starting from a state $s$, we define the state resulting from $s$ by applying a plan $P$, denoted by $s[P]$, inductively as follows. For the empty plan $P = \epsilon$, we define $s[\epsilon] = s$. For non-empty plans $P$ we define $s[P]$ as follows, where $a$ is the last operator in $P$ and $P'$ is the prefix of $P$ up to, but not including, $a$:

- If $\mathsf{pre}(a) \neq s[P']\!\restriction\!\mathsf{vars}(\mathsf{pre}(a))$ (that is, the preconditions of $a$ are not satisfied in the state $s[P']$), then $s[P', a] = s[P']$.

- Otherwise, $s[P', a]$ is the state equal to $\mathsf{post}(a)$ on variables $v \in \mathsf{vars}(\mathsf{post}(a))$, and equal to $s[P']$ on variables $v \in V \setminus \mathsf{vars}(\mathsf{post}(a))$.

A plan $P$ is a *solution plan* if $\mathsf{init}[P]$ is a goal state.

We are concerned with the computational problem *plan existence* (PlanExist): given an instance $\Pi = (V, \mathsf{init}, \mathsf{goal}, A)$, decide if there exists a solution plan.

## 2.2 Graphs

A *directed graph* is a pair $(V, E)$ where $V$ is the *vertex set* and $E \subseteq V \times V$ is the *edge set*. An *undirected graph* is a pair $(V, E)$ where $V$ is the vertex set and $E \subseteq \{\{u, v\} \mid u, v \in V\}$ is the edge set. We will often only say graph and edge if it is clear from the context whether it is directed or undirected. The notation $V(G)$ refers to the vertex set of a graph $G$ and $E(G)$ refers to its edge set. If $e = (u, v)$ or $e = \{u, v\}$ is an edge, then the vertices $u$ and $v$ are *incident* with $e$. Furthermore, the directed edge $(u, v)$ is an *outgoing edge* of $u$ and an *incoming edge* of $v$. For a directed graph $G = (V, E)$, we write $U(G)$ to denote the correspsonding undirected graph $U(G) = (V, E_U)$ where $E_U = \{\{u, v\} \mid (u, v) \in E\}$. That is, $U(G)$ is the undirected graph induced by $G$ by ignoring the orientation of edges.

Let $G = (V, E)$ be a directed graph and let $v_0, \ldots, v_k \in V$ such that $v_1, \ldots, v_k$ are distinct and $(v_{i-1}, v_i) \in E$ for all $i$ $(1 \leq i \leq k)$. Then the sequence $v_0, \ldots, v_k$ is a *directed path* of length $k$ in $G$ if $v_0 \neq v_k$ and it is a *directed cycle* of length $k$ in $G$ if $v_0 = v_k$. Paths and cycles in undirected graphs are defined analogously, except that there is no direction to consider. A graph is *acyclic* if it contains no cycles.

Let $G = (V, E)$ be a directed graph and let $v \in V$ be a vertex. Then, $v$ is *isolated* if it has no incoming or outgoing edges, $v$ is a *source* if it has at least one outgoing edge but no incoming edge, $v$ is a *sink* if it has at least one incoming edge but no outgoing edge and otherwise $v$ is *intermediate*.

Let $G = (V_G, E_G)$ and $H = (V_H, E_H)$ be two directed graphs. Then $G$ and $H$ are *isomorphic* (denoted $G \simeq H$) if there exists a bijective function $f : V_G \to V_H$ such that $(u, v) \in E_G$ if and only if $(f(u), f(v)) \in E_H$. Furthermore, $H$ is a *subgraph* of $G$ if $V_H \subseteq V_G$ and $E_H \subseteq E_G \cap (V_H \times V_H)$. When $E_H = E_G \cap (V_H \times V_H)$ we say that the subgraph $H$ is *induced* by the vertex set $V_H$. Isomorphisms and subgraphs are analogously defined for undirected graphs.

Let $G$ be an undirected graph. Then $G$ is *connected* if there is a path between every pair of vertices in $G$. A *connected component* of $G$ is a maximal subgraph of $G$ that is A Refined View of Causal Graphs and Component SizesA Refined View of Causal Graphs and Component SizesA Refined View of Causal Graphs and Component Sizes connected. Let $G$ be a directed graph. Then $G$ is *weakly connected* if $U(G)$ is connected. A *weakly connected component* of $G$ is a maximal subgraph of $G$ that is weakly connected. That is, in a weakly connected component there are paths between every pair of vertices if we ignore the direction of edges. Let $G = (V_G, E_G)$ and $H = (V_H, E_H)$ be two directed graphs such that $V_G$ and $V_H$ are disjoint. Then the *(disjoint) union* of $G$ and $H$ is defined as $G \cup H = (V_G \cup V_H, E_G \cup E_H)$ and is a commutative operation. Note that if a graph $G$ consists of the (weakly) connected components $G_1, \ldots, G_n$, then $G = G_1 \cup G_2 \cup \ldots \cup G_n$.

We further define some numeric graph parameters. For a directed graph $G$ and a vertex $v \in V(G)$, the *indegree* of $v$ is $|\{u \in V(G) \mid (u,v) \in E(G)\}|$, i.e. the number of incoming edges incident with $v$, and the *outdegree* of $v$ is $|\{u \in V(G) \mid (v,u) \in E(G)\}|$, i.e. the number of outgoing edges incident with $v$. For an undirected graph $G$, the *degree* of $v \in V(G)$ is $|\{u \in V(G) \mid \{v,u\} \in E(G)\}|$, i.e. the number of edges incident with $v$. We extend this to graphs as follows. If $G$ is an undirected graph, then $\mathsf{deg}(G)$ denotes the largest degree of any vertex in $V(G)$. Similarly, if $G$ is a directed graph then $\mathsf{in\text{-}deg}(G)$ denotes the largest indegree of any vertex in $V(G)$ and $\mathsf{out\text{-}deg}(G)$ denotes the largest outdegree of any vertex in $V(G)$. Furthermore, if $G$ is an undirected graph, then $\mathsf{path\text{-}length}(G)$ denotes the length of the longest path in $G$ and $\mathsf{cc\text{-}size}(G)$ denotes the size of the largest connected component in $G$. If $G$ is a directed graph, then $\mathsf{path\text{-}length}(G)$ denotes the length of longest directed path in $G$. We also define $\mathsf{upath\text{-}length}(G) = \mathsf{path\text{-}length}(U(G))$ and $\mathsf{cc\text{-}size}(G) = \mathsf{cc\text{-}size}(U(G))$. That is, $\mathsf{upath\text{-}length}(G)$ is the length of the longest path in $G$ if ignoring the direction of edges and $\mathsf{cc\text{-}size}(G)$ is the size of the largest weakly connected component in $G$. Note that if $G$ is an undirected connected graph, then $\mathsf{path\text{-}length}(G)$ equals the diameter of $G$. We extend all such numeric graph properties ($\mathsf{in\text{-}deg}$, $\mathsf{path\text{-}length}$ etc.) to sets of graphs such that if $\mathcal{C}$ is a set of graphs and $\mathsf{prop}$ is a graph property, then $\mathsf{prop}(\mathcal{C}) = \max_{G \in \mathcal{C}} \mathsf{prop}(G)$.

### 2.3 Special Graph Types

In the literature on causal graphs, as well as in this article, there are certain types of graphs that are of particular interest and that are thus useful to refer to by names. We distinguish the following types of undirected graphs: A *tree* is an undirected graph in which any two vertices are connected by exactly one path, i.e. it is acyclic and connected. A *path graph* is a tree where all vertices have degree 1 or 2, i.e. it is a tree that does not branch. A *star graph* is a tree where all vertices except one, the *centre vertex*, have degree 1.

For directed graphs, we distinguish the following types: An *in-star graph* is a directed graph $G$ such that $U(G)$ is a star graph and all edges are directed towards the centre. An *out-star graph* is a directed graph $G$ such that $U(G)$ is a star graph and all edges are directed out from the centre. A *directed path graph* is a directed graph $G$ such that $U(G)$ is a path graph, $\mathsf{in\text{-}deg}(G) \leq 1$ and $\mathsf{out\text{-}deg}(G) \leq 1$, i.e. $G$ is a directed path over all its vertices and contains no other edges. A *polytree* is a directed graph $G$ such that $U(G)$ is a tree, i.e. $G$ is a weakly connected directed graph that can be constructed from a tree by giving a unique direction to every edge. A *polypath* is a directed graph $G$ such that $U(G)$ is a path graph, i.e. $G$ is a weakly connected directed graph that can be constructed from a path graph by giving a unique direction to every edge. A *fence* is a polypath where every vertex is either a source or a sink, i.e. the edges alternate in direction at every vertex.

It should be noted that the out-star graph is usually called a directed star graph in graph theory, while the in-star graph appears to have no standard name. We hence deviate sligthly from standard terminology in order to have logical names for both graph types. Also the polypath appears to have no standard name, but polypath is a logical term in analogy with polytree. It should be further noted that a parallel terminology for certain graph types has evolved in the literature on causal graphs in planning. For instance, in-stars, out-stars and directed paths are commonly referred to as *inverted forks*, *forks* and *directed chains*, respectively.

Note that the number of sinks and sources in a polypath differ by at most one, i.e. a polypath with $m$ sinks has $m + c$ sources for some $c \in \{-1, 0, 1\}$. Furthermore, every fence is a polypath, but not every polypath is a fence.

We define the following graphs and graphs classes:

- $S_k^{\text{in}}$ denotes the in-star graph with one centre vertex and $k$ sources. Also define the class $\mathbf{S}^{\text{in}} = \{S_k^{\text{in}} \mid k \geq 0\}$.

- $S_k^{\text{out}}$ denotes the out-star with one centre vertex and $k$ sinks. Also define the class $\mathbf{S}^{\text{out}} = \{S_k^{\text{out}} \mid k \geq 0\}$.

- $dP_k$ denotes the directed path on $k$ vertices. Also define the class $\mathbf{dP} = \{dP_k \mid 1 \leq k\}$.

- $F_m^c$, for $c \in \{-1, 0, 1\}$, denotes the fence with $m$ sinks and $m + c$ sources. Also define the class $\mathbf{F}^c = \{F_m^c \mid 1 \leq m\}$, for each $c \in \{-1, 0, 1\}$, and the class $\mathbf{F} = \mathbf{F}^{-1} \cup \mathbf{F}^0 \cup \mathbf{F}^{+1}$.

Examples of these graph types are illustrated in Figure 1.

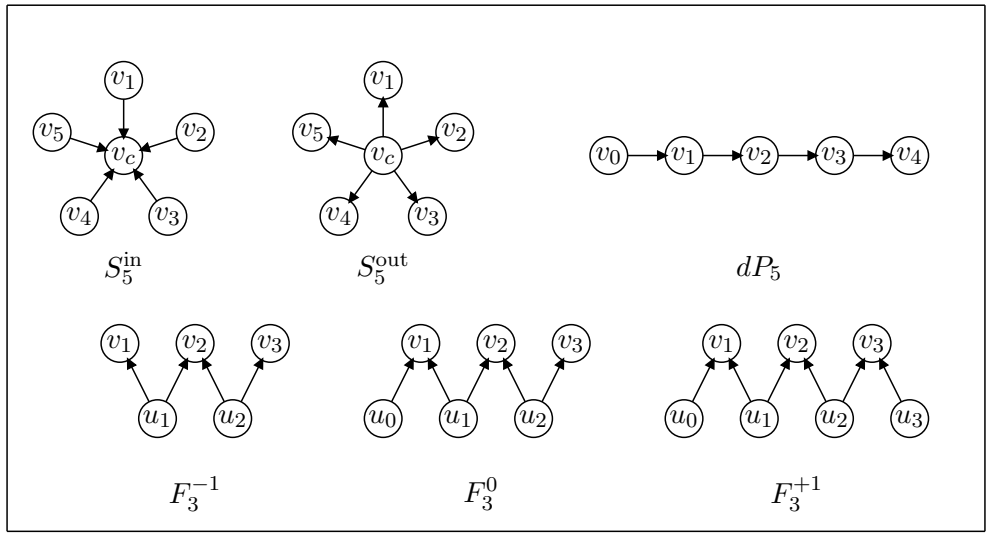

Figure 1: Examples of some important graph types.

The following observation about polypaths will be used later on.

**Proposition 1.** *Let $G$ be a polypath with at most $m$ sinks and $m + 1$ sources such that path-length$(G) \leq k$. Then $|V(G)| \leq 2mk + 1$.*

*Proof.* There are at most $2m$ distinct paths from a source to a sink, each of these having at most $k-1$ intermediate vertices. Hence $|V(G)| \leq m + (m+1) + 2m(k-1) = 2mk + 1$. □

This bound is obviously tight in the case where there are $m$ sinks and $m+1$ sources, and every path from a source to a sink contains exactly $k-1$ intermediate vertices.

## 3. Structurally Restricted Planning

The topic of study in this article is causal graphs for planning, but before discussing this concept we first define the concept of domain-transition graphs (Jonsson & Bäckström, 1998a). Although not used explicitly in any of our results, it is useful for explaining some of the proofs later in the article. Let $\Pi = (V, \text{init}, \text{goal}, A)$ be a planning instance. For each variable $v \in V$, we define the *domain-transition graph (DTG)* for $v$ as a directed graph $(\mathsf{D}(v), E)$, where for all $x, y \in \mathsf{D}(v)$, $E$ contains the edge $(x, y)$ if there is some operator $a \in A$ such that $\text{post}(a)(v) = y$ and either $\text{pre}(a)(v) = x$ or $v \notin \text{vars}(\text{pre}(a))$.

The causal graph for a planning instance describes how the variables of the instance depends on each other, as implicitly defined by the operators.

**Definition 2.** The *causal graph* of a planning instance $\Pi = (V, \text{init}, \text{goal}, A)$ is the directed graph $CG(\Pi) = (V, E)$ where $E$ contains the edge $(u, v)$ for every pair of distinct vertices $u, v \in V$ such that $u \in \text{vars}(\text{pre}(a)) \cup \text{vars}(\text{post}(a))$ and $v \in \text{vars}(\text{post}(a))$ for some operator $a \in A$.

The causal graph gives some, but not all, information about the operators. For instance, if the causal graph is acyclic, then all operators must be unary, i.e. $|\text{vars}(\text{post})(a)| = 1$ for all operators, since any non-unary operator must necessarily introduce a cycle according to the definition. However, the presence of cycles does not necessarily mean that there are non-unary operators. For instance, if both the edges $(u, v)$ and $(v, u)$ are present in the graph, then this can mean that there is some operator $a$ such that both $u \in \text{vars}(\text{post}(a))$ and $v \in \text{vars}(\text{post}(a))$. However, it can also mean that there are two operators $a$ and $a'$ such that $u \in \text{vars}(\text{pre}(a))$, $v \in \text{vars}(\text{post}(a))$, $v \in \text{vars}(\text{pre}(a'))$ and $u \in \text{vars}(\text{post}(a'))$, which could thus both be unary operators. Similarly, the degree of the vertices provides an upper bound on the number of pre- and postconditions of the operators, but no lower bound. Suppose there is a vertex $u$ with indegree 2 and incoming edges $(v, u)$ and $(w, u)$. This could mean that there is some operator $a$ such that $u \in \text{vars}(\text{post}(a))$ and both $v \in \text{vars}(\text{pre}(a))$ and $w \in \text{vars}(\text{pre}(a))$. However, it can also mean that there are two different operators $a$ and $a'$ such that $v \in \text{vars}(\text{pre}(a))$, $u \in \text{vars}(\text{post}(a))$, $w \in \text{vars}(\text{pre}(a'))$ and $u \in \text{vars}(\text{post}(a'))$.

The PlanExist problem is extended from planning instances to causal graphs in the following way. For a class $\mathcal{C}$ of directed graphs, $\mathsf{PlanExist}(\mathcal{C})$ is the problem of deciding for an arbitrary planning instance $\Pi$ such that $CG(\Pi) \in \mathcal{C}$, whether $\Pi$ has a solution or not. That is, the complexity of $\mathsf{PlanExist}(\mathcal{C})$ refers to the complexity of the set of planning instances whose causal graphs are members of $\mathcal{C}$.

There are a number of results in the literature on the computational complexity of planning for various classes of causal graphs. However, these results usually assume that the graph class has a restricted structure, e.g. containing only in-stars or only directed paths. A more general and abstract result is the following theorem.

**Theorem 3.** (Chen & Giménez, 2010, Thm. 3.1) *Let $\mathcal{C}$ be a class of directed graphs. If cc-size($\mathcal{C}$) is bounded, then PlanExist($\mathcal{C}$) is solvable in polynomial time. If cc-size($\mathcal{C}$) is unbounded, then PlanExist($\mathcal{C}$) is not polynomial-time solvable (unless $\boldsymbol{W[1]} \subseteq \boldsymbol{nu\text{-}FPT}$).*

While the theorem describes a crisp borderline between tractable and intractable graph classes, it does so under the assumption that $\mathbf{W}[1] \not\subseteq \mathbf{nu\text{-}FPT}$[1]. Both these complexity classes are from the theory of parameterised complexity and cannot be immediately related to the usual complexity classes. It is out of the scope of this article to treat parameterised complexity and we refer the reader to standard textbooks (Downey & Fellows, 1999; Flum & Grohe, 2006). The result in the theorem is not a parameterised result, however; it is only the condition that is parameterised, so it suffices to note that the intractability result holds under a condition that is difficult to relate to other common assumptions, such as $\mathbf{P} \neq \mathbf{NP}$. One of the reasons why Chen and Giménez were forced to state the theorem in this way was that a classification into polynomial and **NP**-hard classes would not have been exhaustive, since there are graph classes that are **NP**-intermediate. (A problem is **NP**-intermediate if it is neither in **P** nor **NP**-complete, unless $\mathbf{P} = \mathbf{NP}$.)

This theorem might be viewed as the starting point for the research reported in this article, where we investigate this problem from the perspective of standard complexity classes. For instance, **NP**-hardness can be proved in the case of unbounded components if adding further restrictions, which we will do in Section 5.

## 4. Basic Constructions

This section presents some results that are necessary for the theorems later in the article. The first three results, that planning is **NP**-hard for in-stars (aka. inverted forks), out-stars (aka. forks) and directed paths (aka. directed chains), are known from the literature, while the **NP**-hardness result for fences is new. We will, however, provide new proofs also for the in-star and out-star cases. The major reason is that in Section 6 we will need to refer to reductions that have certain precisely known properties. Furthermore, the original proofs are only published in a technical report (Domshlak & Dinitz, 2001b) and may thus be hard to access.

**Lemma 4.** (Domshlak & Dinitz, 2001a, Thm. 3.IV) *PlanExist($\mathbf{S}^{\text{in}}$) is **NP**-hard. This result holds even when restricted to operators with at most 2 preconditions and 1 postcondition.*

*Proof.* (New proof) Proof by reduction from 3SAT to a class of planning instances with causal graphs in $\mathbf{S}^{\text{in}}$. The reduction constructs a planning instance where each source in the causal graph corresponds to one of the variables in the formula and the centre corresponds to the clauses. The construction is illustrated in Figure 2 and formally defined as follows.

Let $F = c_1 \wedge \ldots \wedge c_m$ be an arbitrary 3SAT formula with variables $x_1, \ldots, x_n$ and clauses $c_1, \ldots, c_m$. Construct a corresponding planning instance $\Pi_F = (V, \text{init}, \text{goal}, A)$ as follows:

- $V = \{v_c, v_1, \ldots, v_n\}$, where
  $\mathsf{D}(v_c) = \{0, \ldots, m\}$ and
  $\mathsf{D}(v_i) = \{u, f, t\}$, for all $i$ ($1 \leq i \leq n$).

---

1. The condition can be simplified to $\mathbf{W}[1] \not\subseteq \mathbf{FPT}$ if the class $\mathcal{C}$ is recursively enumerable.

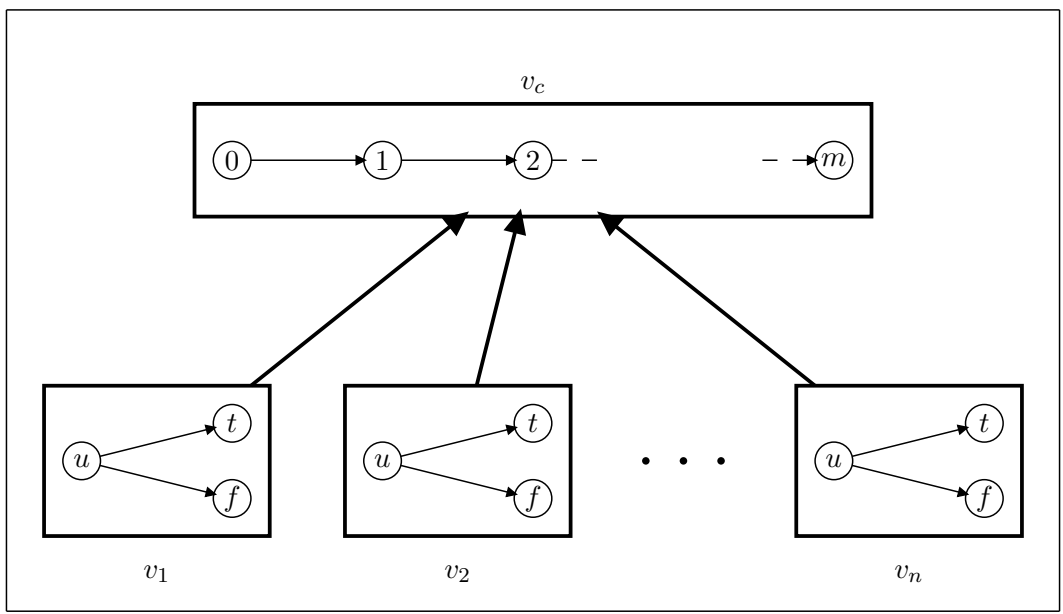

Figure 2: The in-star causal graph and the DTGs for the construction in the proof of Lemma 4.

- $\mathsf{init}(v_i) = u$, for all $i$ $(1 \leq i \leq n)$, and $\mathsf{init}(v_c) = 0$.

- $\mathsf{goal}(v_c) = m$ and $\mathsf{goal}$ is otherwise undefined.

- $A$ consists of the following operators:

  - For each $i$ $(1 \leq i \leq n)$, $A$ contains the operators
    $\mathsf{set\text{-}f}(i) = \langle v_i = u \,;\, v_i = f \rangle$ and
    $\mathsf{set\text{-}t}(i) = \langle v_i = u \,;\, v_i = t \rangle$.
  - For each clause $c_i = (\ell_i^1 \vee \ell_i^2 \vee \ell_i^3)$ and each $j$ $(1 \leq j \leq 3)$, there is some $k$ such that $\ell_i^j = x_k$ or $\ell_i^j = \overline{x_k}$, so let $A$ contain either the operator
    $\mathsf{verify\text{-}clause\text{-}pos}(i,j) = \langle v_c = i-1, v_k = t \,;\, v_c = i \rangle$, if $\ell_i^j = x_k$,
    or the operator
    $\mathsf{verify\text{-}clause\text{-}neg}(i,j) = \langle v_c = i-1, v_k = f \,;\, v_c = i \rangle$, if $\ell_i^j = \overline{x_k}$.

Clearly, the instance $\Pi_F$ can be constructed in polynomial time and $CG(\Pi_F) = S_n^{\mathrm{in}}$, so it remains to prove that $\Pi_F$ has a solution if and only if $F$ is satisfiable.

Each source variable $v_i$ can be changed independently. It starts with the undefined value $u$ and can be set to either $t$ or $f$, corresponding to true and false, respectively, for the corresponding variable $x_i$ in $F$. Once it is set to either $t$ or $f$, it cannot be changed again. That is, variables $v_1, \ldots, v_n$ can be used to choose and commit to a truth assignment for $x_1, \ldots, x_n$. The centre variable $v_c$ has one value, $i$, for each clause $c_i$ in $F$, plus the initial value 0. It is possible to reach the goal value $m$ from the inital value 0 by stepping through all intermediate values in numerical order. For each such step, from $i-1$ to $i$, there are three operators to choose from, corresponding to each of the literals in clause $c_i$. The step is possible only if one of $v_1, \ldots, v_n$ is set to a value consistent with one of the literals in $c_i$. That is, the goal $v_c = m$ can be achieved if and only if variables $v_1, \ldots, v_n$ are set to values corresponding to a truth assignment for $x_1, \ldots, x_n$ that satisfies $F$.

The restricted case (with respect to pre- and post-conditions) is immediate from the construction above. □

The problem is known to be tractable, though, if the domain size of the centre variable is bounded by a constant (Katz & Domshlak, 2010). Furthermore, the causal graph heuristic by Helmert (2004) is based on identifying in-star subgraphs of the causal graph, and it should be noted that he provided a variant of the original proof due to some minor technical differences in the problem formulations.

**Lemma 5.** (Domshlak & Dinitz, 2001a, Thm. 3.III) *PlanExist*($\mathbf{S}^{\text{out}}$) *is* **NP**-*hard. This result holds even when restricted to operators with at most 1 precondition and 1 postcondition.*

*Proof.* (New proof) Proof by reduction from 3SAT to a class of planning instances with causal graphs in $\mathbf{S}^{\text{out}}$. The reduction constructs a planning instance where the centre vertex of the causal graph corresponds to the variables in the formula and each sink corresponds to one of the clauses. The construction is illustrated in Figure 3 and formally defined as follows.

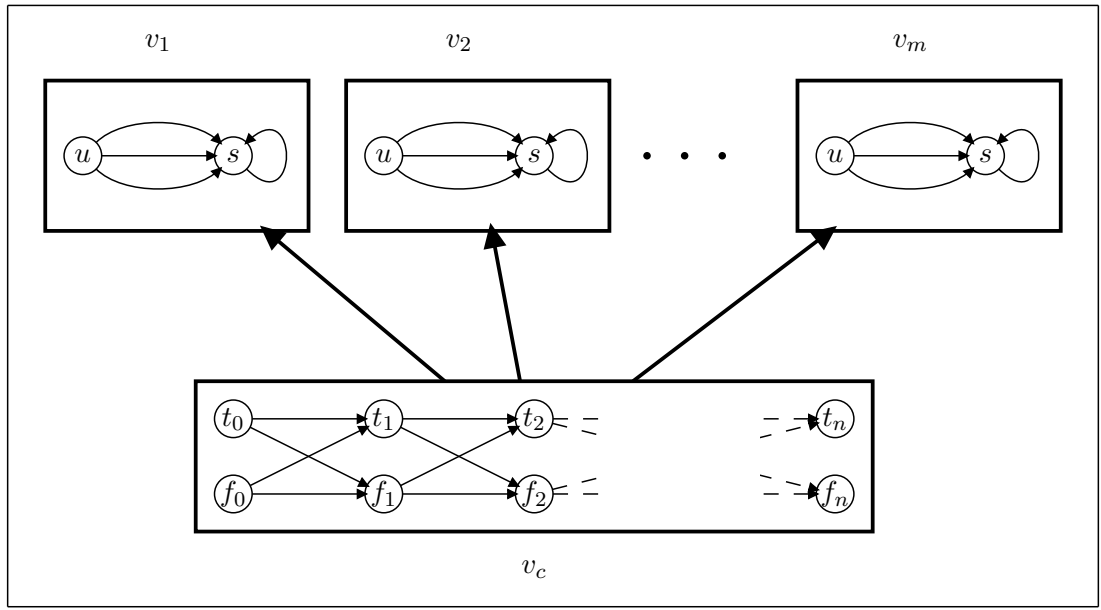

Figure 3: The out-star causal graph and the DTGs for the construction in the proof of Lemma 5.

Let $F = c_1 \wedge \ldots \wedge c_m$ be an arbitrary 3SAT formula with variables $x_1, \ldots, x_n$ and clauses $c_1, \ldots, c_m$. Construct a corresponding planning instance $\Pi_F = (V, \text{init}, \text{goal}, A)$ as follows:

- $V = \{v_c, v_1, \ldots, v_m\}$, where
  $\mathsf{D}(v_c) = \{f_0, \ldots, f_n, t_0, \ldots, t_n\}$ and
  $\mathsf{D}(v_i) = \{u, s\}$, for all $i$ $(1 \leq i \leq m)$.

- $\mathsf{init}(v_i) = u$, for all $i$ $(1 \leq i \leq m)$, and $\mathsf{init}(v_c) = f_0$.

- $\mathsf{goal}(v_i) = s$, for all $i$ $(1 \leq i \leq m)$, and $\mathsf{goal}(v_c)$ is undefined.

- $A$ consists of the following operators:
  - For each $i$ $(1 \leq i \leq n)$, $A$ contains the operators
    step-c$(f_{i-1}, f_i) = \langle v_c = f_{i-1} \, ; \, v_c = f_i \rangle$,
    step-c$(f_{i-1}, t_i) = \langle v_c = f_{i-1} \, ; \, v_c = t_i \rangle$,
    step-c$(t_{i-1}, f_i) = \langle v_c = t_{i-1} \, ; \, v_c = f_i \rangle$ and
    step-c$(t_{i-1}, t_i) = \langle v_c = t_{i-1} \, ; \, v_c = t_i \rangle$.
  - For each clause $c_i = (\ell_i^1 \vee \ell_i^2 \vee \ell_i^3)$ and each $j$ $(1 \leq j \leq 3)$, there is a $k$ such that $\ell_i^j = x_k$ or $\ell_i^j = \overline{x_k}$, so let $A$ contain either the operator
    verify-clause-pos$(i, j) = \langle v_c = t_k \, ; \, v_i = s \rangle$, if $\ell_i^j = x_k$,
    or the operator
    verify-clause-neg$(i, j) = \langle v_c = f_k \, ; \, v_i = s \rangle$, if $\ell_i^j = \overline{x_k}$.

Clearly, the instance $\Pi_F$ can be constructed in polynomial time and $CG(\Pi_F) = S_n^{\mathrm{out}}$, so it remains to prove that $\Pi_F$ has a solution if and only if $F$ is satisfiable.

Variable $v_c$ can be changed independently and it has two values, $t_i$ and $f_i$, for each variable $x_i$ in $F$, corresponding to the possible truth values for $x_i$. In addition there is an initial value $f_0$ (and a dummy value $t_0$ in order to simplify the formal definition). Both the values $t_n$ and $f_n$ are reachable from the initial value $f_0$, and each such plan will correspond to a path $f_0, z_1, z_2, \ldots, z_n$ where each $z_i$ is either $t_i$ or $f_i$. That is, $v_c$ must pass either value $t_i$ or $f_i$, but not both, for each $i$. Hence, any such path will correspond to a truth assignment for the variables $x_1, \ldots, x_n$ in $F$. For each clause $c_i$ in $F$, there is a corresponding variable $v_i$ that can change value from the initial value $u$, unsatisfied, to the goal value $s$, satisfied. Each $v_i$ has three operators to do this, one for each literal in $c_i$. That is, if $c_i$ contains a literal $x_k$ (or $\overline{x_k}$) then $v_i$ can change value from $u$ to $s$ while $v_c$ has value $t_k$ (or $f_k$). Hence, the goal $v_1 = \ldots = v_m = s$ can be achieved if and only if there is a path for $v_c$ that corresponds to a truth assignment for $x_1, \ldots, x_n$ that satisfies $F$. (Note, though, that $v_c$ must not always follow a path all the way to $f_n$ or $t_n$ since a partial assignment may sometimes be sufficient to prove satisfiability.)

The restricted case (with respect to pre- and post-conditions) is immediate from the construction above. □

The problem is known to be tractable, though, if the domain size of the centre variable is bounded by a constant (Katz & Keyder, 2012).

The following result on planning with directed-path causal graphs is also known from the literature.

**Lemma 6.** (Giménez & Jonsson, 2009, Prop. 5.5) *PlanExist(**dP**) is **NP**-hard, even when all variables have domain size 5 and the operators have at most 2 preconditions and 1 post-condition.*

We refer to Giménez and Jonsson for the proof. However, we will implicitly use their proof later in this article so there are a few important observations to make about it. The reduction is from SAT and, thus, works also as a reduction from 3SAT. Furthermore, the reduction transforms a formula with $n$ variables and $m$ clauses to a planning instance with $(2m+4)n$ variables. As a final remark, this problem is known to be tractable if all variables have a domain of size 2 (Domshlak & Dinitz, 2001a).

While the three previous results are known in the literature, the following result is new to the best of our knowledge.

**Lemma 7.** *PlanExist*($\mathbf{F}^{+1}$) *is **NP**-hard. This result holds even when restricted to operators with at most 2 preconditions and 1 postcondition.*

*Proof.* Proof by reduction from 3SAT to a class of planning instances with causal graphs in $\mathbf{F}^{+1}$.

The reduction constructs a planning instance where each sink of the causal graph corresponds to one of the clauses in the formula, while each source corresponds to all variables. Furthermore, the source variables are synchronized to have the same behaviour. The construction is illustrated in Figure 4 and formally defined as follows.

Let $F = c_1 \wedge \ldots \wedge c_m$ be an arbitrary 3SAT formula with variables $x_1, \ldots, x_n$ and clauses $c_1, \ldots, c_m$. Construct a corresponding planning instance $\Pi_F$ as follows:

- $V = \{u_0, \ldots, u_m, v_1, \ldots, v_m\}$, where
  $\mathsf{D}(u_i) = \{f_0, \ldots, f_n, t_0, \ldots, t_n\}$, for all $i$ ($0 \leq i \leq m$), and
  $\mathsf{D}(v_i) = \{f_0^u, \ldots, f_m^u, t_0^u, \ldots, t_m^u, f_0^s, \ldots f_m^s, t_0^s, \ldots, t_m^s, s\}$, for all $i$ ($1 \leq i \leq m$).

- $\mathsf{init}(u_i) = f_0$, for all $i$ ($0 \leq i \leq m$), and $\mathsf{init}(v_i) = f_0^u$, for all $i$ ($1 \leq i \leq m$).

- $\mathsf{goal}(v_i) = s$, for all $i$ ($1 \leq i \leq m$), and $\mathsf{goal}$ is otherwise undefined.

- Let $A$ consist of the following operators:

  - For all $i, j$ ($1 \leq i \leq n, 0 \leq j \leq m$), $A$ contains the operators
    step-x$(j, f_{i-1}, f_i) = \langle u_j = f_{i-1} \,;\, u_j = f_i \rangle$,
    step-x$(j, f_{i-1}, t_i) = \langle u_j = f_{i-1} \,;\, u_j = t_i \rangle$,
    step-x$(j, t_{i-1}, f_i) = \langle u_j = t_{i-1} \,;\, u_j = f_i \rangle$ and
    step-x$(j, t_{i-1}, t_i) = \langle u_j = t_{i-1} \,;\, u_j = t_i \rangle$.

  - For all $i, j$, ($1 \leq i \leq n, 1 \leq j \leq m$), $A$ contains the operators
    step-clause-u$(j, f_{i-1}^u, f_i^u) = \langle v_j = f_{i-1}^u, u_{j-1} = f_i, u_j = f_i \,;\, v_j = f_i^u \rangle$,
    step-clause-u$(j, f_{i-1}^u, t_i^u) = \langle v_j = f_{i-1}^u, u_{j-1} = t_i, u_j = t_i \,;\, v_j = t_i^u \rangle$,
    step-clause-u$(j, t_{i-1}^u, f_i^u) = \langle v_j = t_{i-1}^u, u_{j-1} = f_i, u_j = f_i \,;\, v_j = f_i^u \rangle$,
    step-clause-u$(j, t_{i-1}^u, t_i^u) = \langle v_j = t_{i-1}^u, u_{j-1} = t_i, u_j = t_i \,;\, v_j = t_i^u \rangle$,
    step-clause-s$(j, f_{i-1}^s, f_i^s) = \langle v_j = f_{i-1}^s, u_{j-1} = f_i, u_j = f_i \,;\, v_j = f_i^s \rangle$,
    step-clause-s$(j, f_{i-1}^s, t_i^s) = \langle v_j = f_{i-1}^s, u_{j-1} = t_i, u_j = t_i \,;\, v_j = t_i^s \rangle$,
    step-clause-s$(j, t_{i-1}^s, f_i^s) = \langle v_j = t_{i-1}^s, u_{j-1} = f_i, u_j = f_i \,;\, v_j = f_i^s \rangle$,
    step-clause-s$(j, t_{i-1}^s, t_i^s) = \langle v_j = t_{i-1}^s, u_{j-1} = t_i, u_j = t_i \,;\, v_j = t_i^s \rangle$,

  - For each $j$ ($1 \leq j \leq m$), $A$ contains the operators
    finalize-clause-f$(j) = \langle v_j = f_n^s \,;\, v_j = s \rangle$ and
    finalize-clause-t$(j) = \langle v_j = t_n^s \,;\, v_j = s \rangle$.

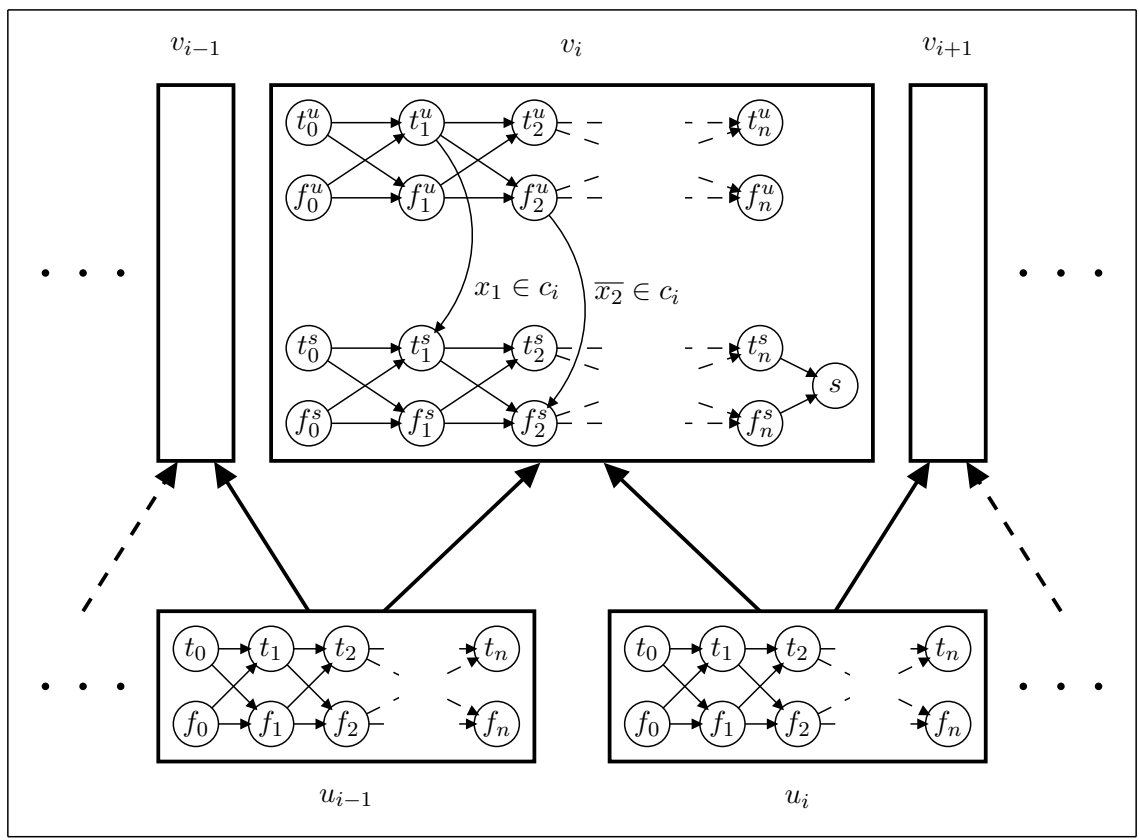

Figure 4: The fence causal graph and the DTGs for the construction in the proof of Lemma 7. (This example assumes that clause $c_i$ contains the literals $x_1$ and $\overline{x_2}$).

- For each clause $c_i = (\ell_i^1 \vee \ell_i^2 \vee \ell_i^3)$ and for each $j$ $(1 \leq j \leq 3)$, there is a $k$ such that $\ell_i^j = x_k$ or $\ell_i^j = \overline{x_k}$ so let $A$ contain either the operator
  verify-pos$(i,j) = \langle v_i = t_k^u \; ; \; v_i = t_k^s \rangle$, if $\ell_i^j = x_k$,
  or the operator
  verify-neg$(i,j) = \langle v_i = f_k^u \; ; \; v_i = f_k^s \rangle$, if $\ell_i^j = \overline{x_k}$.

Clearly, the instance $\Pi_F$ can be constructed in polynomial time and $CG(\Pi_F) = F_m^{+1}$. Hence, it remains to prove that $\Pi_F$ has a solution if and only if $F$ is satisfiable.

First consider only variables $u_i$ and $v_i$, for some $i$. The construction of the domain and the operators for $u_i$ is identical to the one for $v_c$ in the proof of Lemma 5, i.e. there is a directed path from value $f_0$ to $f_n$ or $t_n$ for every possible truth assignment for the variables $x_1, \ldots, x_n$ in $F$. Variable $v_i$, corresponds to clause $c_i$ and contains two copies of the DTG for $u_i$, where the values differ only in the extra superscript, $u$ or $s$. The latter copy is extended with the additional value $s$, denoting that the clause has been satisfied. There are operators that allows $v_i$ to mimic the behaviour of $u_i$; it can follow the corresponding path in either of its two copies. Furthermore, for each of the three literals in $c_i$ there is an operator that makes it possible to move from value $z_k^u$ to value $z_k^s$ if value $z_k$ of $u_i$ is consistent with this literal. Since $v_i$ starts at $f_0^u$ and must reach either $f_m^s$ or $t_m^s$ in order to reach the goal value $s$, it is necessary for $v_i$ to make such a transition for one of the literals in $c_i$. That is, if $u_i$ follows the path $f_0, z_1, \ldots, z_n$ then $v_i$ must follow the path $f_0^u, z_1^u, \ldots, z_k^u, z_k^s, \ldots, z_n^s, s$, for some $k$ such that $x_k$ occurs in a literal in $c_i$ and $z_k$ is a satisfying truth value for this literal.

Now consider also variable $u_{i-1}$. Since each operator that affects the value of $v_i$ either has the same precondition on both $u_{i-1}$ and $u_i$ or no precondition on either, it follows that $u_{i-1}$ and $u_i$ must both choose the same path if $v_i$ is to reach its goal. Since every variable $v_j$ forces synchronization of its adjacent variables $u_{j-1}$ and $u_j$ in this manner, it follows that all of $u_0, \ldots, u_m$ must choose exactly the same path for any plan that is a solution. It thus follows from this and from the argument for $u_i$ and $v_i$ that the goal $v_1 = \ldots = v_m = s$ can be achieved if and only if there is a path that all of $u_0, \ldots, u_m$ can choose such that this path corresponds to a satisfying truth assignment for $F$.

For the restriction, we first note that it is immediate from the construction that operators with 3 preconditions and 1 postcondition are sufficient. To see that 2 preconditions are sufficient, consider the following variation on the construction. Each step-clause-u and step-clause-t operator is replaced with two operators as follows. As an example, consider an operator step-clause-u$(j, f_{i-1}^u, t_i^u)$. First introduce an extra value $ft_i^u$ in $\mathsf{D}(v_j)$. Then replace the operator with two new operators

step-clause-u$(j, f_{i-1}^u, ft_i^u) = \langle v_j = f_{i-1}^u, u_{j-1} = t_i \; ; \; v_j = ft_i^u \rangle$ and
step-clause-u$(j, ft_i^u, t_i^u) = \langle v_j = ft_i^u, u_j = t_i \; ; \; v_j = t_i^u \rangle$.

Consider the step in the DTG for $v_j$ from $f_{i-1}^u$ to $t_i^u$. In the original construction, this is done by the single operator step-clause-u$(j, f_{i-1}^u, t_i^u)$, which requires that both $u_{j-1}$ and $u_j$ have value $t_i$. The modified construction instead requires two steps, first a step from $f_{i-1}^u$ to the new intermediate value $ft_i^u$ and then a step from this value to $t_i^u$. The previous conjunctive constraint that $u_{j-1} = u_j = t_i$ is replaced by a sequential constraint that first $u_{j-1} = t_i$ and then $u_j = t_i$. Although it is technically possible for $u_{j-1}$ to have moved on to a new value when the second step is taken, this does not matter; both $u_{j-1}$ and $u_j$ must still choose exactly the same path in their respective DTGs. $\square$

**Corollary 8.** *PlanExist*$(\mathbf{F}^{-1})$, *PlanExist*$(\mathbf{F}^0)$ *and PlanExist*$(\mathbf{F})$ *are* **NP***-hard.*

*Proof.* Neither of the two outer source vertices, $u_0$ and $u_m$, are necessary in the construction in the previous proof. Hence, by omitting either or both of these the reduction works also for $\mathbf{F}^{-1}$ and $\mathbf{F}^0$. Finally, PlanExist$(\mathbf{F})$ is **NP**-hard since $\mathbf{F}^{+1} \subseteq \mathbf{F}$. $\square$

We now have all the basic results necessary for the main theorems of the following two sections.

## 5. Graph Classes and Closure Properties

Like most other results in the literature, the results in the previous section are about classes consisting of some particular graph type, like the class $\mathbf{S}^{\text{in}}$ of all in-stars or the class $\mathbf{F}$ of all fences. This section will depart from this and instead study graph classes with certain closure properties. We will first discuss the standard concepts of subgraph closure and minor closure, finding that the first does not contain all the graphs we need while the latter results in a set with too many graphs. For that reason, we will define a new concept, SP-closure, which is incomparable with subgraph closure but is a subset of minor closure. We will then show that this closure concept defines a borderline between the non-**NP**-hard graph classes and large number of useful **NP**-hard classes.

### 5.1 Subgraph Closure and Minor Closure

Suppose $\mathcal{C}$ is a class of graphs which is closed under taking subgraphs. Then for every graph $G$ in $\mathcal{C}$ it is the case that every subgraph $H$ of $G$ must also be in $\mathcal{C}$. Subgraph closure is not sufficient for our purposes, though. For instance, a subgraph of a polypath will always be either a polypath or a graph where every weakly connected component is a polypath. However, a polypath need not have any subgraphs that are fences of more than trivial size. We will need a closure property that guarantees that if $\mathcal{C}$ contains a polypath with $m$ sinks, then it also contains a fence with $m$ sinks. An obvious candidate for this is the concept of minor-closure, which is a superset of the subgraph-closure. The concepts of graph minors and minor-closure has rapidly evolved into a very important and useful research area in mathematical as well as computational graph theory (Lovász, 2005; Mohar, 2006).

In order to define graph minors we first need the concept of edge contraction, which is commonly defined as follows, although other definitions occur in the literature.

**Definition 9.** Let $G = (V, E)$ be a directed graph and let $e = (u, v) \in E$ be an edge such that $u \neq v$. Then the *contraction* of $e$ in $G$ results in a new graph $G' = (V', E')$, such that

- $V' = (V \setminus \{u, v\}) \cup \{w\}$ and
- $E' = \{(f(x), f(y)) \mid (x, y) \in E,\ (x, y) \neq (u, v) \text{ and } (x, y) \neq (v, u)\}$,

where $w$ is a new vertex, not in $V$, and the function $f : V \to V'$ is defined such that $f(u) = f(v) = w$ and otherwise $f(x) = x$.

That is, when an edge $(u, v)$ is contracted, the two vertices $u$ and $v$ are replaced with a single new vertex $w$ and all edges that were previously incident with either $u$ or $v$ are redirected to be incident with $w$. Figure 5 shows an example of edge contraction. We say that a graph $H$ is a contraction of another graph $G$ if $H$ can result from contracting zero or more edges in $G$.

The concept of graph minors can now be defined as follows.

**Definition 10.** A directed graph $H$ is a *minor* of a directed graph $G$ if $H$ is isomorphic to a graph that can be obtained by zero or more edge contractions of a subgraph of $G$.

An example is illustrated in Figure 6. The graph $G$ in the figure is a weakly connected directed graph, which also happens to be a polypath. If vertex $v_9$ is removed from $G$, then the restriction to the remaining vertices is still a weakly connected graph which is a subgraph of $G$. Removing also $v_4$ results in the graph $H$, which consists of two weakly connected components $H_1$ and $H_2$. All of $H$, $H_1$ and $H_2$ are subgraphs of $G$, but they are also minors of $G$, since a subgraph is a minor, by definition. Contracting the edge $(v_1, v_2)$ in $H_1$ results in the graph $M_1$, where $w_1$ is the new vertex replacing $v_1$ and $v_2$. Similarly, contracting the edge $(v_8, v_7)$ in $H_2$ results in $M_2$. The graph $M_1$ is a minor of $G$ since it is the result of an edge contraction in the subgraph $H_1$ of $G$ and the graph $M_2$ is analogously a minor of $G$ too. Also the graph $M$, consisting of the two components $M_1$ and $M_2$ is a minor of $G$, since it is the result of two contractions in the subgraph $H$ of $G$. While the graphs $H$, $H_1$ and $H_2$ are both subgraphs and minors of $G$, the graphs $M$, $M_1$ and $M_2$ are only minors of $G$, not subgraphs.

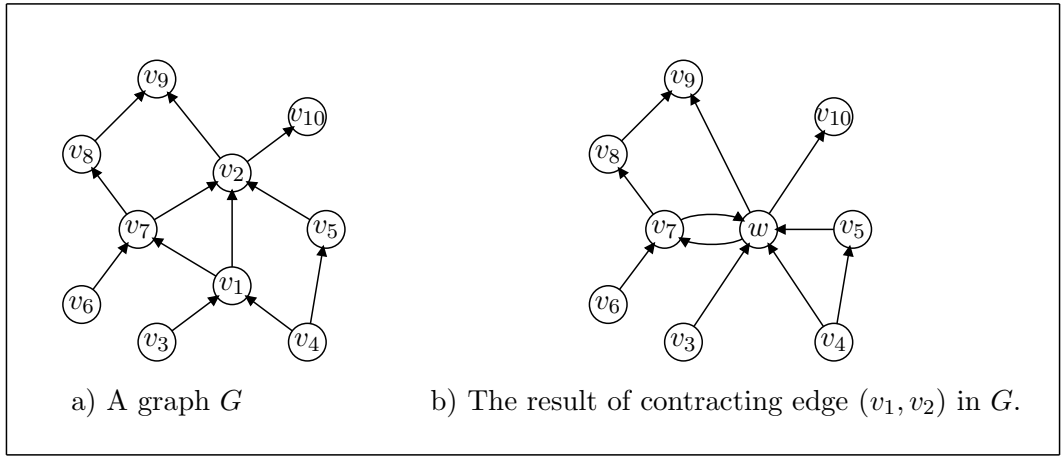

Figure 5: Edge contraction.

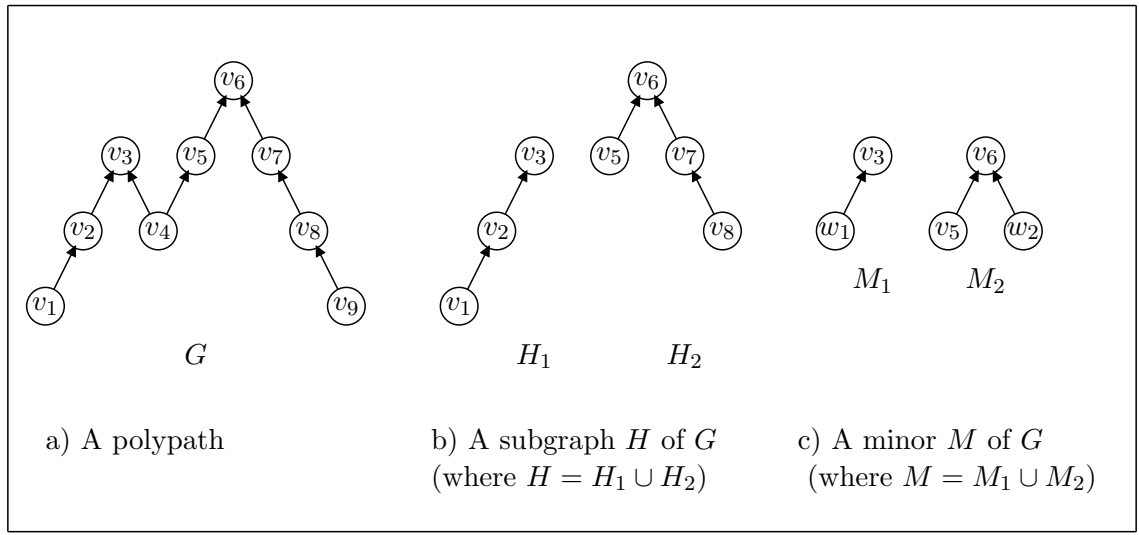

Figure 6: Subgraphs and minors.

A trivial example of a minor-closed class is the class of all graphs, which is minor-closed since it contains all graphs and every minor of a graph is itself a graph. More interestingly, many commonly studied graph types result in minor-closed classes. For instance, the class $\mathbf{S}^{\text{in}}$ of all in-stars is minor-closed, as is the class $\mathbf{S}^{\text{out}}$ of all out-stars and the class $\mathbf{dP}$ of all directed paths. Furthermore, a weakly connected minor of a polypath is a polypath and a weakly connected minor of a polytree is a polytree. As an illustration, once again consider Figure 6. The graph $G$ is a polypath, and the weakly connected graphs $H_1$, $H_2$, $M_1$ and $M_2$ are all minors of $G$, but they are also polypaths. In fact, $M_1$ and $M_2$ are also fences. Note though, that neither $H$ nor $M$ is a polypath, since they both consist of more than one weakly connected component. It is worth noting, however, that the class **F** of all fences is not minor-closed although every fence is a polypath; a weakly connected minor of a fence must be a polypath, but it is not necessarily a fence.

Requiring minor-closed graph classes is, however, overly strong. For instance, it would be sufficient to require that for every graph $G \in \mathcal{C}$, also every weakly connected minor of $G$ is in $\mathcal{C}$. That is, in the example in Figure 6 we would require that $H_1$, $H_2$, $M_1$ and $M_2$ are all in $\mathcal{C}$ if $G$ is in $\mathcal{C}$, but we would not require that also $H$ and $M$ are in $\mathcal{C}$. This is both reasonable and desirable in the context of causal graphs. If the causal graph of a planning instance consists of two or more weakly connected components, then these components correspond to entirely independent subinstances that can be solved separately.

Furthermore, certain natural restrictions do not mix well with minor-closed classes. Consider, for instance, the example in Figure 7, with an acyclic graph $G = (V, E)$, where $V = \{v_1, v_2, v_3, v_4\}$ and $E = \{(v_1, v_2), (v_2, v_3), (v_3, v_4), (v_1, v_4)\}$. If we contract the edge $(v_1, v_4)$ to a new vertex $w$ we get a cycle graph on the vertices $w, v_2, v_3$. That is, a class of acyclic graphs is not minor-closed in general, which is problematic considering the importance of acyclic causal graphs.

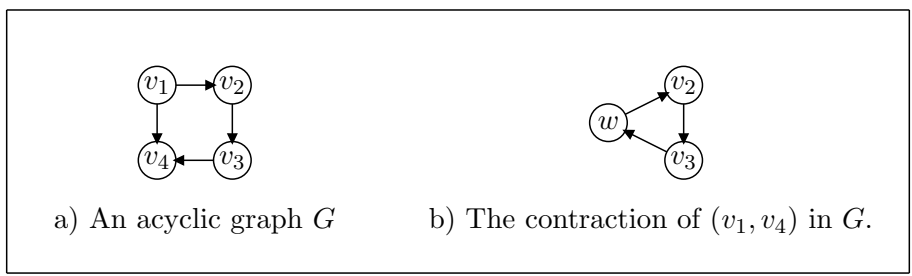

a) An acyclic graph $G$     b) The contraction of $(v_1, v_4)$ in $G$.

Figure 7: Contracting an edge in an acyclic graph can result in a cycle.

### 5.2 SP-Closed Graph Classes

In order to avoid problems with acyclicity (and other similar problems) and to avoid defining special variants of the contraction and minor concepts, we instead identify a set of minimal requirements that a closure must satisfy in order to imply **NP**-hardness for the PlanExist problem. We will focus on one such set of restrictions, defining a concept we refer to as SP-closure (where SP denotes that the set is closed under stars and polypaths).

**Definition 11.** Let $G$ and $H$ be two directed graphs. Then $H$ is an *SP-graph* of $G$ if $H$ is weakly connected and either of the following holds:

1. $H$ is an in-star that is a subgraph of $G$,

2. $H$ is an out-star that is a subgraph of $G$ or

3. $H$ can be obtained by zero or more contractions of some polypath $G'$ such that $G'$ is a subgraph of $G$.

A class $\mathcal{C}$ of graphs is *SP-closed* if it contains every SP-graph of every graph $G \in \mathcal{C}$.

SP-closure has a number of interesting properties, including the following:

**Proposition 12.** *Let $G$ and $H$ be directed graphs and let $\mathcal{C}$ be a class of directed graphs.*

1. *If $G$ is a polypath, then every SP-graph of $G$ is a polypath.*

2. *Every SP-graph of $G$ is acyclic.*

3. *If $H$ is an SP-graph of $G$, then $H$ is a minor of $G$.*

4. *If $\mathcal{C}$ is minor-closed, then $\mathcal{C}$ is SP-closed.*

*Proof.* 1) Suppose $G$ is a polypath. Obviously, $G$ cannot contain an in-star or out-star with higher degree than two, and any such star is also a polypath. Hence, we only need to consider the third case in the definition. We note that any weakly connected subgraph $G'$ of $G$ must also be a polypath, and that doing contractions on a polypath results in a polypath.

2) Immediate since in-stars, out-stars and polypaths are all acyclic and contracting edges cannot introduce a cycle in any of these cases.

3) Immediate from the definitions of minors and SP-graphs.

4) Immediate from 3. □

This proposition says that it makes sense to talk about SP-closed classes of polypaths and SP-closed classes of acyclic graphs. It also says that SP-closure and minor-closure are comparable concepts; the SP-closure of a class is a subset of the minor-closure of the same class.

We can now prove the following result about SP-closed classes of polypaths, which we need for the main theorem.

**Lemma 13.** *Let $\mathcal{C}$ be an SP-closed class of polypaths. If cc-size($\mathcal{C}$) is unbounded, then PlanExist($\mathcal{C}$) is **NP**-hard. This result holds even when restricted to operators with at most 2 preconditions and 1 postcondition.*

*Proof.* Proof by cases depending on whether the directed path length of $\mathcal{C}$ is bounded or not.

*Case 1:* Suppose that path-length($\mathcal{C}$) is unbounded. Let $n > 1$ be an arbitrary integer. Then there must be some graph $G \in \mathcal{C}$ such that $G$ contains a subgraph $H$ that is a directed path graph and $V(H) = n$. Obviously, $H$ is an SP graph of $G$, since a directed path is also a polypath. It follows that $H \in \mathcal{C}$ since $\mathcal{C}$ is SP-closed. Furthermore, $H \simeq dP_n$ so **NP**-hardness of PlanExist($\mathcal{C}$) follows from Lemma 6, since $n$ was choosen arbitrarily.

*Case 2:* Instead suppose that path-length($\mathcal{C}$) $\leq k$ for some constant $k \geq 0$. Let $n > 1$ be an arbitrary integer. Since all graphs in $\mathcal{C}$ are polypaths and cc-size($\mathcal{C}$) is unbounded, there must be some polypath $G \in \mathcal{C}$ such that $V(G) \geq n$. It thus follows from the assumption and Proposition 1 that $G$ must have at least $m$ sinks and $m + 1$ sources, for some $m$ such A Refined View of Causal Graphs and Component Sizesthat $V(G) \leq 2mk+1$. There must, thus, be some subgraph $G'$ of $G$ that is a polypath with exactly $m$ sinks and $m+1$ sources (i.e. $G'$ is weakly connected) and there must, thus, also be a graph $H$ that can be obtained by zero or more contractions of $G'$ such that $H \simeq F_m^{+1}$. It follows that $H \in \mathcal{C}$ since $\mathcal{C}$ is SP-closed. **NP**-hardness of PlanExist($\mathcal{C}$) thus follows from Lemma 7, since $n$ was choosen arbitrarily and $k$ is constant.

To see that the result holds even if the operators under consideration have at most 2 preconditions and 1 postcondition, simply note that this restriction holds for all reductions used in the underlying **NP**-hardness proofs in Section 4. □

Chen and Giménez (2010, Thm. 3.19) proved a similar result: If $\mathcal{C}$ is a class of polypaths[2] with unbounded components and unbounded number of sources, then PlanExist($\mathcal{C}$) is not polynomial-time solvable unless **W**[1] $\subseteq$ **nu-FPT**.

In order to prove the main result of this section, we also need the *Moore bound* (Biggs, 1993, p. 180), which is stated as follows: for an arbitrary connected undirected graph $G$, the maximum number of vertices is

$$|V(G)| \leq 1 + d \sum_{i=0}^{k-1}(d-1)^i, \qquad (1)$$

where $d = \deg(G)$ and $k = \mathsf{path\text{-}length}(G)$.

We can now prove that under the additional restriction that graph classes are SP-closed, we can avoid **NP**-intermediate problems and prove **NP**-hardness for graph classes with unbounded components.

**Theorem 14.** *Let $\mathcal{C}$ be an SP-closed class of directed graphs. If cc-size($\mathcal{C}$) is unbounded, then PlanExist($\mathcal{C}$) is **NP**-hard. This result holds even when restricted to operators with at most 2 preconditions and 1 postcondition and all graphs in $\mathcal{C}$ are acyclic.*

*Proof.* First suppose there is some constant $k$ such that in-deg($\mathcal{C}$) $\leq k$, out-deg($\mathcal{C}$) $\leq k$ and upath-length($\mathcal{C}$) $\leq k$. Consider an arbitrary graph $G \in \mathcal{C}$. Obviously, $\deg(U(G)) \leq 2k$ and path-length($U(G)$) $\leq k$, so it follows from the Moore bound that no component in $U(G)$ can have more than $1 + 2k\sum_{i=0}^{k-1}(2k-1)^i$ vertices. However, since cc-size($G$) = cc-size($U(G)$) and $G$ was choosen arbitrarily, it follows that cc-size($\mathcal{C}$) is bounded. This contradicts the assumption so at least one of in-deg($\mathcal{C}$), out-deg($\mathcal{C}$) and upath-length($\mathcal{C}$) is unbounded. The remainder of the proof is by these three (possibly overlapping) cases.

*Case 1:* Suppose that in-deg($\mathcal{C}$) is unbounded. Let $n > 0$ be an arbitrary integer. Then there must be some graph $G \in \mathcal{C}$ containing a vertex with indegree $n$ or more, so there must also be a subgraph $H$ of $G$ such that $H \simeq S_n^{\mathrm{in}}$. Hence, $H \in \mathcal{C}$ since $\mathcal{C}$ is SP-closed. It thus follows from Lemma 4 that PlanExist($\mathcal{C}$) is **NP**-hard, since $n$ was choosen arbitrarily.

*Case 2:* Suppose that out-deg($\mathcal{C}$) is unbounded. This case is analogous to the previous one, but using Lemma 5 instead of Lemma 4.

*Case 3:* Suppose that upath-length($\mathcal{C}$) is unbounded. Let $n > 0$ be an arbitrary integer. Then there must be some graph $G \in \mathcal{C}$ such that $U(G)$ contains a path of length $n$, and there must, thus, also be a subgraph $H$ of $G$ such that $H$ is a polypath of length $n$. Obviously, $H$

---

2. Chen and Giménez use the term *source-sink configuration* for polypath.

595

is an SP-graph of $G$ (doing zero contractions) so $H \in \mathcal{C}$ since $\mathcal{C}$ is SP-closed. It thus follows from Lemma 13 that PlanExist($\mathcal{C}$) is **NP**-hard, since $n$ was choosen arbitrarily.

To see that the result holds even if the operators under consideration have at most 2 preconditions and 1 postcondition, simply note that this restriction holds for all reductions used in the underlying **NP**-hardness proofs in Section 4. Similarly, the acyclicity restriction holds since the result is based only on in-stars, out-stars and polypaths, which are all acyclic graphs. □

This theorem is somewhat more restricted than the one by Chen and Giménez since it requires the additional constraint that $\mathcal{C}$ is SP-closed. On the other hand, it demonstrates that SP-closure is a sufficient condition to avoid graph classes such that PlanExist is **NP**-intermediate and, thus, sharpen the result to **NP**-hardness. It should be noted, though, that this is not an exact characterization of all graph classes that are **NP**-hard for PlanExist. There are other such graph classes, but SP-closure captures a large number of interesting graph classes. For instance, the class of all acyclic graphs is SP-closed (recall that this class is not minor-closed), although not every subclass of it is SP-closed. As an opposite example, any non-empty class that does not contain a single acyclic graph cannot be SP-closed.

## 6. Beyond SP-Closed Graph Classes

This section is divided into three parts. We first discuss why the previous results, as well as most other similar **NP**-hardness results in the literature, are problematic, which motivates us to switch over to non-uniform complexity theory. The second part contains a number of preparatory results that are required for the main theorem in the third part.

### 6.1 Why NP-Hardness is Not Enough

We refer to a planning problem as *generic* if it has instances of varying size, depending on one or more parameters. An archetypical example is the blocks world, where the natural parameter is the number of blocks. For a particular encoding and a specified number of blocks, the variables and operators will be the same whatever the inital state and goal is. That is, if we fix the encoding then we get a *planning frame* $\Phi_n = (V_n, A_n)$ for every number, $n$, of blocks. That is, $\Phi_n$ is the same for all instances with $n$ blocks and is thus a function of $n$. All instances $(V_n, \text{init}, \text{goal}, A_n)$ with $n$ blocks will be instantiations of $\Phi_n$ with different init and goal components but with the same $V_n$ and $A_n$ components. An instance can thus be specified with three unique parameters, $n$, init and goal, where only the first parameter, $n$, affects the size of the instance. Furthermore, the causal graph for an instance depends only on the variables and the operators, which means that all instantiations of a frame $\Phi_n$ have the same causal graph, which we denote $CG(\Phi_n)$. The class of causal graphs for blocks world instances will be $\mathcal{D} = \{CG(\Phi_1), CG(\Phi_2), CG(\Phi_3), \ldots\}$, although $\Phi_1, \Phi_2, \Phi_3, \ldots$, and thus also $\mathcal{D}$, will differ depending on the encoding.

It is often possible to analyse the complexity of a particular generic planning problem. Examples of this are the complexity of blocks-world planning (Gupta & Nau, 1992) and the complexity of various problems from the International Planning Competitions (IPC) (Helmert, 2003, 2006b). In the context of this article, though, we are rather interested in the complexity of the class of causal graphs corresponding to a generic problem, than the complexity of the specific problem itself. Suppose that a class $\mathcal{D}$ of causal graphs happens to be a subset of some class $\mathcal{C}$ of graphs such that we know that PlanExist($\mathcal{C}$) is tractable. Then we can infer that also PlanExist($\mathcal{D}$) is tractable, and thus also that all generic planning problems with causal graphs in $\mathcal{D}$ are tractable. However, in order to prove that PlanExist($\mathcal{D}$) is **NP**-hard (or hard for some other complexity class) we would have to prove that there is some class $\mathcal{C}$ of graphs such that PlanExist($\mathcal{C}$) is **NP**-hard and $\mathcal{C}$ is a subset of $\mathcal{D}$. Finding such a class $\mathcal{C}$ may not be trivial, though.

One problem is that the encoding can have a large influence on how densely or sparsely the causal graphs occur with respect to size. Consider, for instance, blocks world encodings with multi-valued variables and with boolean variables respectively. A typical encoding with multi-valued variables will use one variable for the status of the hand and two variables for each block, one for the position of the block and one to flag whether the block is clear or not. That is, such encodings will use $2n + 1$ variables for an $n$-block frame. An encoding with boolean variables, on the other hand, will typically represent the block position with a number of boolean variables, one for each other block that a block can be on. A boolean encoding will thus use $n^2 + 1$ variables for an $n$-block frame. While $\mathcal{D}$ will contain a graph for every odd number of vertices in the first case, it will be increasingly sparse in the second case. The class $\mathcal{D}$ of causal graphs for a generic planning problem will, thus, typically not be SP-closed, or even closed under taking subsets. Furthermore, since $\mathcal{D}$ will typically not contain a member for every possible number of vertices, it cannot possibly contain any of the known **NP**-hard sets $\mathbf{S}^{\text{in}}$, $\mathbf{S}^{\text{out}}$, $\mathbf{dP}$ etc. as a subset. Hence, in order to prove that a class $\mathcal{D}$ of causal graphs is hard for **NP** (or some other complexity class), it will often be necessary to make a dedicated proof for $\mathcal{D}$. This is often doable, however. A generic planning problem has a corresponding function $f$ that takes a parameter value $n$, e.g. the number of blocks in blocks world, such that $f(n) = \Phi_n$. If $f$ is furthermore polynomial-time computable in the value of $n$, which will often be the case, then also the corresponding causal graph, $CG(\Phi_n)$, is polynomial-time computable. However, even if this can be done for many generic planning problems, it will be a specific proof for every specific encoding of every particular generic planning problem. The same holds for particular classes of causal graphs; every specific class will typically require its own dedicated proof.

In order to get around these problems and to be able to prove a more general result that does not depend on the specific planning problems or causal graphs, we switch over to non-uniform complexity. This makes it possible to prove more powerful results, while retaining natural connections with the ordinary complexity classes. The basic vehicle for proving non-uniform complexity results is the *advice-taking Turing machine*, which is defined as follows.

**Definition 15.** An *advice-taking Turing machine* $M$ has an associated sequence of *advice strings* $A_0, A_1, A_2, \ldots$, a special advice tape and an *advice function* $A$, from the natural numbers to the advice sequence, s.t. $A(n) = A_n$. On input $x$ the advice tape is immediately loaded with $A(||x||)$. After that $M$ continues like an ordinary Turing machine, except that it also has access to the advice written on the advice tape.

If there exists a polynomial $p$ s.t. $||A(n)|| \leq p(n)$, for all $n > 0$, then $M$ is said to use *polynomial advice*. The complexity class **P**/poly is the set of all decision problems that can be solved on some advice-taking TM that runs in polynomial time using polynomial advice.

Note that the advice depends only on the size of the input, not its content, and need not even be computable. Somewhat simplistically, an advice-taking Turing machine is a machine that has an infinite data-base with constant access time. However, for each input size there is only a polynomial amount of information while there might be an exponential number of instances sharing this information. The power of polynomial advice is thus still somewhat limited and useful relationships are known about how the non-uniform complexity classes relate to the standard ones are known. One such result is the following.

**Theorem 16.** (Karp & Lipton, 1980, Thm. 6.1) *If $\mathbf{NP} \subseteq \mathbf{P}/poly$, then the polynomial hierarchy collapses to the second level.*

### 6.2 Preparatory Results

Before carrying on to the main theorem of this section, we need a few auxiliary results. We first show that if a planning instance has a causal graph $G$ that is a subgraph of some graph $H$, then the instance can be extended to an equivalent instance with $H$ as causal graph.

**Lemma 17.** *Let $\Pi$ be a planning instance and let $G$ be a directed graph such that $CG(\Pi)$ is a subgraph of $G$. Then there is a planning instance $\Pi_G$ such that*

- *$\Pi_G$ can be constructed from $\Pi$ in polynomial time,*
- *$CG(\Pi_G) = G$ and*
- *$\Pi_G$ has a solution if and only if $\Pi$ has a solution.*

*Furthermore, $\Pi_G$ has the same maximum number of pre- and postconditions for its operators as $\Pi$ (or one more if this value is zero in $\Pi$).*

*Proof.* Let $\Pi = (V, \mathsf{init}, \mathsf{goal}, A)$ be a planning instance and let $CG(\Pi) = (V, E)$. Let $G = (V_G, E_G)$ be a directed graph such that $CG(\Pi)$ is a subgraph of $G$. Let $U = V_G \setminus V$. Construct a planning instance $\Pi_G = (V_G, \mathsf{init}_G, \mathsf{goal}_G, A_G)$ as follows:

- $\mathsf{D}_G(u) = \{0, 1\}$, for all $u \in U$, and
  $\mathsf{D}_G(v) = \mathsf{D}(v) \cup \{\star\}$, for all $v \in V$, (where $\star$ is a new value not in $\mathsf{D}(v)$).

- $\mathsf{init}_G(v) = \mathsf{init}(v)$, for all $v \in V$, and
  $\mathsf{init}_G(u) = 0$, for all $u \in U$.

- $\mathsf{goal}_G(v) = \mathsf{goal}(v)$, for all $v \in V$, and
  $\mathsf{goal}_G(u)$ is undefined for all $u \in U$.

- Let $A_G$ consist of the following operators:

    - Let $A_G$ contain all $a \in A$.
    - For each edge $(x, v) \in E_G \setminus E$ such that $x \in V_G$ and $v \in V$, let $A_G$ also contain an operator $\mathsf{star}(x, v) = \langle x = 0 \,;\, v = \star \rangle$.
    - For each edge $(x, u) \in E_G$ such that $x \in V_G$ and $u \in U$, let $A_G$ also contain an operator $\mathsf{set}(x, u) = \langle x = \mathsf{init}(x) \,;\, u = 1 \rangle$.

Obviously $\Pi_G$ can be constructed in polynomial time and $CG(\Pi_G) = G$, so it remains to prove that $\Pi_G$ has a solution if and only if $\Pi$ has a solution.

Suppose $P = a_1, \ldots, a_n$ is a plan for $\Pi$. Then $P$ is also a plan for $\Pi_G$ since $\mathsf{goal}_G(u)$ is undefined for all $u \in U$ and $a_1, \ldots, a_n \in A_G$. To the contrary, suppose $P = a_1, \ldots, a_n$ is a plan for $\Pi_G$. For each operator $a_i$ in $P$, there are three cases: (1) $a_i \in A$, (2) $a_i$ is a set operator or (3) $a_i$ is a star operator. In case 2, operator $a_i$ serves no purpose since it only modifies some variable in $U$, which has an undefined goal value. In case 3, operator $a_i$ sets some variable $v \in V$ to $\star$ and has no effect on any other variables. If $\mathsf{goal}_G(v)$ is undefined, then $a_i$ serves no purpose. Otherwise there must be some operator $a_j$, $j > i$, such that $a_j$ can change $v$ from $\star$ to some value in $\mathsf{D}(v)$, i.e. $a_i$ serves no purpose in this case either. It follows that the operator sequence $P'$ obtained from $P$ by removing all operators that are not in $A$ is also a plan for $\Pi_G$. Furthermore, since $P'$ contains only operators from $A$ it is also a plan for $\Pi$. It follows that $\Pi$ has a plan if and only if $\Pi_G$ has a plan. □

This construction increases the maximum domain size by one but has very little effect on the maximum number of pre- and postconditions. This is suitable for our purpose, since we do not consider the influence of domain sizes in this article. Other constructions are possible if we want to balance the various factors differently.

In the proof of the forthcoming theorem we will also do the opposite of taking graph minors, that is, starting from a minor $G$ of some target graph $H$ we will extend $G$ to $H$. In order to do so, we need an operation similar to the opposite of edge contraction. This is satisfied by a graph operation known as *edge subdivision*.

**Definition 18.** Let $G = (V, E)$ be a directed graph and let $(u, v) \in E$ be an edge such that $u \neq v$. Then the *subdivision* of $(u, v)$ in $G$ is a graph $G' = (V \cup \{w\}, E')$ where $w$ is a new vertex and $E' = (E \setminus \{(u, v)\}) \cup \{(u, w), (w, v)\}$.

Although one might consider other definitions, e.g. in the case where both $(u, v)$ and $(v, u)$ are in $E$, this one is sufficient for our purpose and it follows the usual extension to directed graphs (cf., Kühn, Osthus, & Young, 2008). Usually an operation called *smoothing* is considered as the inverse of edge subdivision. However, smoothing can be viewed as a restricted case of edge contraction, so it is reasonable to think of edge subdivision as a sort of inverse of edge contraction. An example of edge subdivision is illustrated in Figure 8.
We further note that just like an edge contraction of a polypath is a polypath, also an edge subdivision of a polypath is a polypath.

We also need an operation on planning instances corresponding to edge subdivision in their causal graphs. For that purpose, we need a concept of *variable substitution* for operators. We denote the substitution of a variable $w$ for a variable $v$ in a partial state $s$ with $a[v/w]$, defined as:

$$s[v/w](x) = \begin{cases} s(v), & \text{if } x = w, \\ s(x), & \text{if } x \in \mathsf{vars}(s) \setminus \{v, w\}, \\ \text{undefined}, & \text{otherwise.} \end{cases}$$

If $a$ is an operator, then the operator $a' = a[v/w]$ is defined such that $\mathsf{pre}(a') = \mathsf{pre}(a)[v/w]$ and $\mathsf{post}(a') = \mathsf{post}(a)[v/w]$.

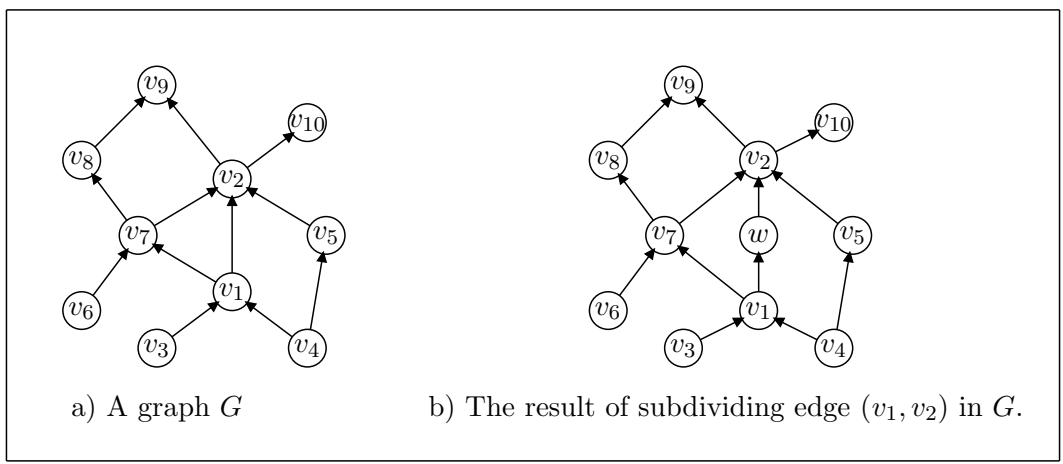

Figure 8: Edge subdivision.

We now have the necessary concepts for modifying an arbitrary planning instance such that the result corresponds to subdividing an edge in the causal graph of the instance. However, we will only need to do this for instances where the causal graph is a polypath. Before proving that this can be done, we first need the following lemma, which states a certain reordering property for plans when the causal graph is a polypath. If we choose an arbitrary vertex $v$ in a polypath $G$ and remove $v$ from $G$, then $G$ falls apart into two weakly connected components $C_1$ and $C_2$. In other words, the vertices of $G$ can be partitioned into three sets $C_0$, $C_1$ and $C_2$ such that $C_0 = \{v\}$ and there is no edge directly between a vertex in $C_1$ and a vertex in $C_2$. It then follows from the definition of causal graphs that no operator that changes some variable in $C_1$ can have a precondition on a variable in $C_2$ and vice versa. The following lemma utilises this fact to prove that any sequence of operators that does not change variable $v$ can be reordered such that all operators that change variables in $C_1$ come before all operators that change variables in $C_2$.

**Lemma 19.** *Let $\Pi = (V, \text{init}, \text{goal}, A)$ be a planning instance such that $G = CG(\Pi)$ is a polypath. Let $v$ be an arbitrary variable in $V$, let $C_0 = \{v\}$ and let $C_1, C_2 \subseteq V$ be the two (possibly empty) weakly connected components of $G$ that result if vertex $v$ is removed from $G$. Define $A_i = \{a \in A \mid \text{vars}(\text{post}(a)) \subseteq C_i\}$ for all $i$ ($0 \leq i \leq 2$). Let $P$ be a plan for $\Pi$. Let $P_1$, $P_2$ and $Q$ be operator sequences such that $P = P_1, Q, P_2$ and $Q$ contains no operator from $A_0$. Let $Q_1$ be the subsequence of $Q$ containing only operators from $A_1$ and let $Q_2$ be the subsequence of $Q$ containing only operators from $A_2$. Then $P_1, Q_1, Q_2, P_2$ is a plan for $\Pi$.*

*Proof.* Assume $C_0$, $C_1$ and $C_2$ as defined in the lemma and recall that $C_0 = \{v\}$. First note that $G$ is acyclic since it is a polypath, so all operators in $A$ are unary. It follows that $\{A_0, A_1, A_2\}$ is a partition of $A$ and, thus, that $A_0 \cup A_1 \cup A_2 = A$. Let $s_0 = \text{init}[P_1]$. Obviously, $(\text{vars}(\text{pre}(a)) \cap C_2 = (\text{vars}(\text{post}(a)) \cap C_2 = \varnothing$ for all $a$ in $Q_1$ and $(\text{vars}(\text{pre}(a)) \cap C_1 = (\text{vars}(\text{post}(a)) \cap C_1 = \varnothing$ for all $a$ in $Q_2$, i.e. for any state $s$ it holds that $s[a] \restriction C_2 = s \restriction C_2$ for all $a$ in $Q_1$ and that $s[a] \restriction C_1 = s \restriction C_1$ for all $a$ in $Q_2$. Furthermore, for any state $s$ it holds that $s[a](v) = s(v)$ for all $a$ in $Q$, since $a \notin A_0$. It follows that $s_0[Q] \upharpoonright C_1 = s_0[Q_1] \upharpoonright C_1$ and $s_0[Q] \upharpoonright C_2 = s_0[Q_2] \upharpoonright C_2$. Hence,

$s_0[Q_1, Q_2] \upharpoonright C_0 = s_0[Q] \upharpoonright C_0$,
$s_0[Q_1, Q_2] \upharpoonright C_1 = s_0[Q_1] \upharpoonright C_1 = s_0[Q] \upharpoonright C_1$ and
$s_0[Q_1, Q_2] \upharpoonright C_2 = s_0[Q_2] \upharpoonright C_2 = s_0[Q] \upharpoonright C_2$.

That is, $s_0[Q_1, Q_2] = s_0[Q]$ and it follows that also $P_1, Q_1, Q_2, P_2$ is a plan for $\Pi$. □

We now prove that if $\Pi$ is a planning instance such that $CG(\Pi)$ is a polypath, then we can subdivide any edge in $CG(\Pi)$ and create a planning instance $\Pi'$ such that $CG(\Pi')$ is this subdivision of $CG(\Pi)$ and $\Pi'$ is solvable if and only if $\Pi$ is solvable.

**Lemma 20.** *Let $\Pi$ be a planning instance such that $CG(\Pi)$ is a polypath and let $e$ be an edge in $CG(\Pi)$. Then there is a planning instance $\Pi'$ such that*

- $\Pi'$ *can be constructed from $\Pi$ in polynomial time,*
- $CG(\Pi')$ *is an edge subdivision of $e$ in $CG(\Pi)$ and*
- $\Pi'$ *has a solution if and only if $\Pi$ has a solution.*

*Proof.* Let $\Pi = (V, \mathsf{init}, \mathsf{goal}, A)$ be a planning instance such that $CG(\Pi)$ is a polypath and let $e = (u, v)$ be an edge in $CG(\Pi)$. Construct a new instance $\Pi' = (V', \mathsf{init}', \mathsf{goal}', A')$ as follows:

- $V' = V \cup \{w\}$, where $\mathsf{D}(w) = \mathsf{D}(u)$ and $w \notin V$.

- $\mathsf{init}'(v) = \mathsf{init}(v)$, for all $v \in V$, and
  $\mathsf{init}'(w) = \mathsf{init}(u)$.

- $\mathsf{goal}' = \mathsf{goal}$.

- Let $A'$ consist of the following groups of operators:

  1. Let $A'$ contain all operators $a \in A$ such that $u \notin \mathsf{vars}(\mathsf{pre}(a))$ or $v \notin \mathsf{vars}(\mathsf{post}(a))$.
  2. Let $A'$ contain the operator $a[u/w]$ for every operator $a \in A$ such that $u \in \mathsf{vars}(\mathsf{pre}(a))$ and $v \in \mathsf{vars}(\mathsf{post}(a))$.
  3. Let $A'$ contain an operator $\mathsf{copy}(u, w, x) = \langle u = x\ ;\ w = x\rangle$ for every value $x \in \mathsf{D}(v)$.

The operators in group 1 are the original operators from $A$ corresponding to all edges in $CG(\Pi)$ except $(u, v)$. The operators in group 2 are the operators from $A$ corresponding to edge $(u, v)$ but modified to instead correspond to the new edge $(w, v)$. The operators in group 3 correspond to the new edge $(u, w)$ and are defined such that variable $w$ can 'mimic' variable $u$. Clearly, this is a polynomial-time construction and $CG(\Pi')$ is an edge subdivision of $CG(\Pi)$. It remains to prove that $\Pi'$ has a plan if and only if $\Pi$ has a plan.

*If:* Suppose $P = a_1, \ldots, a_n$ is a plan for $\Pi$. Construct a new operator sequence $P'$ over $A'$ from $P$ as follows: First, for each $a_i$ in $P$ such that $u \in \mathsf{vars}(\mathsf{pre}(a_i))$ and $v \in \mathsf{vars}(\mathsf{post}(a_i))$, replace $a_i$ with $a_i[u/w]$. Then, for each $a_i$ in $P$ such that $u \in \mathsf{vars}(\mathsf{post}(a_i))$, let $x = \mathsf{post}(a_i)(u)$ and add operator $\mathsf{copy}(u, w, x)$ between $a_i$ and $a_{i+1}$. The resulting sequence $P'$ is a plan for $\Pi'$.

*Only if:* Suppose $P = a_1, \ldots, a_n$ is a plan for $\Pi'$. Define the corresponding state sequence $s_0, \ldots, s_n$ such that $s_0 = \mathsf{init}'$ and $s_i = s_0[a_1, \ldots, a_i]$ for all $i$ ($1 \leq i \leq n$). Without losing generality, assume that $P$ is a shortest plan for $\Pi'$, which implies that $a_i$ is applicable in $s_{i-1}$ for every $i$ ($1 \leq i \leq n$). Define three variable sets $C_0$, $C_1$ and $C_2$ as in Lemma 19 such that $C_0 = \{w\}$, $v \in C_1$ and $u \in C_2$. Also define the corrsponding partition $\{A_0, A_1, A_2\}$ of $A'$, i.e. $A_i = \{a \in A' \mid \mathsf{vars}(\mathsf{post}(a)) \subseteq C_i\}$ for all $i$ ($0 \leq i \leq 2$). Then $A_0$ contains all copy operators and nothing else. Before proving the main result of this direction, we first prove the following auxiliary result:

According to Lemma 19 we can assume that every longest subsequence $a_k, \ldots, a_\ell$ that does not contain any operator from $A_0$ is on the form $a_k, \ldots, a_m, a_{m+1}, \ldots, a_\ell$ such that $a_k, \ldots, a_m \in A_1$ and $a_{m+1}, \ldots, a_\ell \in A_2$. Since it is a longest such sequence, it must hold that either (1) $k = 1$ or (2) $a_{k-1} \in A_0$. In case (1) we have $s_{k-1} = s_0 = \mathsf{init}'$, so $s_{k-1}(u) = s_{k-1}(w)$ since $\mathsf{init}'(u) = \mathsf{init}'(w)$. In case (2) operator $a_{k-1} = \mathsf{copy}(u, w, x)$ for some $x$ such that $s_{k-1}(w) = s_{k-2}(u) = x$. Hence, $s_{k-1}(u) = s_{k-1}(w) = x$ since $a_{k-1}$ does not change $u$. That is, in either case we have $s_{k-1}(u) = s_{k-1}(w)$. Furthermore, for all $i$ ($k \leq i \leq m$) it holds that $s_i \restriction (C_0 \cup C_2) = s_{k-1} \restriction (C_0 \cup C_2)$ since $a_i \in A_1$. It follows that $s_i(u) = s_i(w)$ for all $i$ ($k \leq i \leq m$). Now, for every $i$ ($k \leq i \leq \ell$), if $w \in \mathsf{vars}(\mathsf{pre}(a_i))$ then $a_i$ must be on the form $a[u/w]$, for some $a \in A$, so $v \in \mathsf{vars}(\mathsf{pre}(a_i))$ by definition. Hence, $a_i \in A_1$ so $i \leq m$ and it follows that $s_{i-1}(u) = s_{i-1}(w)$. Since this proof holds for all longest subsequences not containing any operator from $A_0$ we can conclude the following, which will be used below:

(*) For any operator $a_i$ in $P$ such that $a_i = a[u/w]$ for some $a \in A$, it holds that $s_{i-1}(u) = s_{i-1}(w)$.

We now prove the main result of this direction, that also $\Pi$ has a plan since $\Pi'$ has a plan. We do so by constructing a plan $P''$ for $\Pi$ from $P$ in two steps. First we construct an intermediate operator sequence $P'$ and then construct the plan $P''$ from $P'$. The sequence $P'$ is technically not a plan for either $\Pi$ or $\Pi'$, but this intermediate step makes the proof clearer. Temporarily introduce a virtual dummy operator $\mathsf{dum}$ that has no precondition and no postcondition, i.e. it is applicable in any state and has no effect. Then construct the new operator sequence $P' = b_1, \ldots, b_n$ over $A \cup \{\mathsf{dum}\}$ as follows:

- If $a_i \in A$, then $b_i = a_i$.
- If $a_i$ is a copy operator, then $b_i = \mathsf{dum}$.
- Otherwise, $a_i = a[u/w]$ for some operator $a \in A$, so let $b_i$ be that operator $a$.

Define the corresponding state sequence $t_0, \ldots, t_n$ such that $t_0 = \mathsf{init}'$ and $t_i = t_0[b_1, \ldots, b_i]$ for all $i$ ($1 \leq i \leq n$). We claim that $t_i \restriction V = s_i \restriction V$ for all $i$ ($0 \leq i \leq n$). Proof by induction over $i$:

*Basis:* $t_0 = s_0$ by definition.

*Induction:* Suppose $t_{i-1} \restriction V = s_{i-1} \restriction V$ for some $i$ ($1 \leq i \leq n$). There are three cases:

(1) $a_i = b_i$ and $a_i \in A$. Then $w$ is not in the pre- or postcondition of either $a_i$ or $b_i$ so $b_i$ is applicable in $t_{i-1}$ since $a_i$ is applicable in $s_{i-1}$ and $t_{i-1} \restriction V = s_{i-1} \restriction V$ by assumption. Furthermore, $t_i \restriction V = t_{i-1}[b_i] \restriction V = s_{i-1}[a_i] \restriction V = s_i \restriction V$.

(2) $a_i$ is a copy operator and $b_i = \mathsf{dum}$. It is immediate from the definition of $b_i$ that it is applicable in $t_{i-1}$ and that $t_i = t_{i-1}$. Furthermore, $\mathsf{vars}(\mathsf{post}(a_i)) \cap V = \varnothing$ so $s_i \!\restriction\! V = s_{i-1} \!\restriction\! V$. Since $t_{i-1} \!\restriction\! V = s_{i-1} \!\restriction\! V$ by assumption it thus follows that $t_i \!\restriction\! V = s_i \!\restriction\! V$.

(3) $a_i$ is $b_i[u/w]$ and $b_i \in A$. It follows from (*) that $s_{i-1}(w) = s_{i-1}(u)$, so $s_{i-1}(w) = t_{i-1}(u)$ since $u \in V$ and $t_{i-1} \!\restriction\! V = s_{i-1} \!\restriction\! V$ by assumption. Since $a_i$ is applicable in $s_{i-1}$, $\mathsf{pre}(a_i)(w) = \mathsf{pre}(b_i)(u)$ and $\mathsf{pre}(a_i)(x) = \mathsf{pre}(b_i)(x)$ for all variables in $V \setminus \{u\}$, it follows that $b_i$ is applicable in $t_{i-1}$. By definition, $\mathsf{vars}(\mathsf{post}(b_i)) = \mathsf{vars}(\mathsf{post}(a_i)) = \{v\}$, since both $a_i$ and $b_i$ must be unary, and it thus also follows from the definition that $\mathsf{post}(b_i) = \mathsf{post}(a_i)$. Hence, it also follows that $t_i \!\restriction\! V = s_i \!\restriction\! V$, since $t_{i-1} \!\restriction\! V = s_{i-1} \!\restriction\! V$ by assumption.

We have thus shown that $t_i \!\restriction\! V = s_i \!\restriction\! V$ for all $i$ ($0 \leq i \leq n$). Furthermore, clearly $t_i = t_{i-1}$ for all $i$ such that $b_i = \mathsf{dum}$. It follows that we can create a plan $P''$ for $\Pi$ by removing all dummy operators from $P'$.

We conclude that $\Pi$ has a solution if and only if $\Pi'$ has a solution. $\square$

We will finally need the following observations about 3SAT instances. Let $F$ be a 3SAT formula with $n$ variables and $m$ clauses. If it contains no repeated clauses, then

$$\frac{n}{3} \leq m \leq 8n^3 \text{ and, thus, } \left(\frac{m}{8}\right)^{1/3} \leq n \leq 3m.$$

Furthermore, $F$ can be represented as a list of $3m$ literals which requires $3m(1 + \log n) \leq 3m(1 + \log 3m)$ bits, plus some overhead. Hence, $F$ can be represented by at most $cm^2$ bits, for some constant $c$, and we will later use the upper bound $40m^3$, which is safe.

We also note that the reduction used in the proof of Lemma 6 transforms a 3SAT instance with $n$ variables and $m$ clauses to a planning instance with $N = (2m + 4)n$ variables. However, $n \leq 3m$ so $N \leq (2m + 4) \cdot 3m = 6m^2 + 12m$, which can be safely overestimated with $N \leq 18m^2$.

### 6.3 The Main Theorem

We are now prepared to state and prove the main theorem of this section. It follows from the proof of Theorem 14 that if $\mathsf{in\text{-}deg}(\mathcal{C})$, $\mathsf{out\text{-}deg}(\mathcal{C})$ and $\mathsf{upath\text{-}length}(\mathcal{C})$ are all bounded for a class $\mathcal{C}$ of graphs, then $\mathsf{cc\text{-}size}(\mathcal{C})$ is bounded. In that case it is immediate from Theorem 3 that planning is tractable for $\mathcal{C}$. This begs the question what happens if these parameters are not bounded by a constant, yet bounded by some slow-growing function? We will consider the case when they are allowed to grow slowly, as long as they are polynomially related to the instance size. Since we have also noted that practical planning problems will typically not have a causal graph of every size, we will only require that for every graph $G$ in $\mathcal{C}$ there must also be some larger graph $G'$ in $\mathcal{C}$ of size at most $p(|G|)$, for some polynomial $p$. We also define the parameter $\tau(G) = \max\{\mathsf{upath\text{-}length}(G), \mathsf{in\text{-}deg}(G), \mathsf{out\text{-}deg}(G)\}$, and require that $\tau(G)$ and $\|G\|$ are polynomially related. It turns out that planning is still hard under these restrictions, as the following theorem says.

**Theorem 21.** *Let $p$ and $q$ be increasing polynomials on the natural numbers. Let $\mathcal{C}$ be a class of directed graphs containing a subset of weakly connected graphs $G_1, G_2, G_3, \ldots$ such that:*

1. $|V(G_1)| \leq p(q(1))$,
   $|V(G_{i-1})| < |V(G_i)| \leq p(|V(G_{i-1})|)$, *for all $i > 1$, and*

603

2. $|V(G_i)| \leq q(\tau(G_i))$, for all $i \geq 1$.

*If PlanExist($\mathcal{C}$) is polynomial-time solvable, then the polynomial hierarchy collapses. This result holds even when restricted to operators with at most 2 preconditions and 1 postcondition and all graphs in $\mathcal{C}$ are acyclic.*

*Proof.* Let $G_1, G_2, G_3, \ldots$ be a sequence of weakly connected graphs in $\mathcal{C}$ as assumed in the theorem. Let $H_1, H_2, H_3, \ldots$ be a sequence of graphs defined as follows: for each $i > 0$, $H_i = G_j$ for the smallest $j$ such that $q(i) \leq |V(G_j)|$.

We first prove that $i$ underestimates $\tau(H_i)$. Combining the requirement that $q(i) \leq |V(G_j)|$ with condition 2 of the theorem, that $|V(G_j)| \leq q(\tau(G_j))$, we get $q(i) \leq |V(G_j)| \leq q(\tau(G_j))$. Since $H_i = G_j$ we get $q(i) \leq |V(H_i)| \leq q(\tau(H_i))$, that is, that $i \leq \tau(H_i)$. It follows that also $i \leq |V(H_i)|$ holds.

We then prove that $|V(H_i)|$ is polynomially bounded by $p(q(i))$. Since $j$ is choosen as the smallest value satisfying that $q(i) \leq |V(G_j)|$, it must be that either $j = 1$ or $|V(G_{j-1})| < q(i)$. If $j = 1$, then $H_i = G_j = G_1$ and $|V(G_1)| \leq p(q(1))$ by condition 1 in the theorem. Hence, $|V(H_i)| = |V(G_1)| \leq p(q(1)) \leq p(q(i))$, since $p$ and $q$ are increasing. Otherwise, when $j > 1$, condition 1 of the lemma says that $|V(G_j)| \leq p(|V(G_{j-1})|)$. Combining this with the inequality $|V(G_{j-1})| < q(i)$ yields that $|V(G_j)| \leq p(|V(G_{j-1})|) < p(q(i))$, that is, $|V(H_i)| \leq p(q(i))$ since $H_i = G_j$. Combining this with the previous result that $i \leq |V(H_i)|$ and the construction of $H_i$ yields that $H_1, H_2, H_3$ is a sequence of graphs with non-decreasing and unbounded size.

Now, define a sequence $A_0, A_1, A_2, \ldots$ of tuples such that for all $i \geq 0$, either of the following holds:

1. in-deg($H_i$) $\geq i$ and $A_i = $ (in-deg, $H_i, X_i$) such that $X_i$ is a subgraph of $H_i$ and $X_i \simeq S_i^{\text{in}}$.

2. out-deg($H_i$) $\geq i$ and $A_i = $ (out-deg, $H_i, X_i$) such that $X_i$ is a subgraph of $H_i$ and $X_i \simeq S_i^{\text{out}}$.

3. upath-length($H_i$) $\geq i$ and $A_i = $ (upath-length, $H_i, X_i$) such that $X_i$ is a subgraph of $H_i$ and $X_i$ is a polypath of length $i$.

For every $i > 0$, at least one of these three cases must hold since $i \leq \tau(H_i)$.

Define an advice-taking Turing machine $M$ that uses the sequence $A_1, A_2, A_3, \ldots$ as advice and takes 3SAT formulae as input. Assume that the representation of each formula $F$ is padded to size $40m^3$ bits, where $m$ is the number of clauses. Although somewhat redundant, this is still a reasonable encoding in the sense of Garey and Johnson (1979). Let $M$ work as follows. Let $F$ be an input formula with $n$ variables and $m$ clauses and let $t = ||F|| = 40m^3$. Then the advice is $A_t = (\mathsf{x}, H_t, X_t)$. First $M$ constructs a planning instance $\Pi_F$. There are three cases depending on $\mathsf{x}$:

**$\mathsf{x} = $ in-deg:** By construction, $X_t$ is a subgraph of $H_t$ such that $H_t \simeq S_t^{\text{in}}$. Since $t = 40m^3$ and $n \leq 3m$, it follows that $n \leq t$, so $X_t$ contains a subgraph $H'$ such that $H' \simeq S_n^{\text{in}}$. Construct $\Pi_F$ in the same way as in the proof of Lemma 4, using the vertices of $H'$ as variables. Then, $CG(\Pi_F) = H'$.

**$\mathsf{x} = $ out-deg:** Analogous to previous case, but constructing $\Pi_F$ according to the proof of Lemma 5 instead.

**x = upath-length:** By construction, $X_t$ is a subgraph of $H_t$ which is a polypath of length $t = 40m^3$. Suppose that $X_t$ contains less than $m$ sinks and $m+1$ sources and that path-length$(X_t) < 18m^2$. It then follows from Proposition 1 that

$$|V(X_t)| \;<\; 2m \cdot 18m^2 + 1 \;=\; 36m^3 + 1 \;<\; 40m^3 \;=\; t.$$

However, this contradicts the construction so $X_t$ must either contain a directed path of length $18m^2$ or have at least $m$ sinks and $m+1$ sources.

1. If $X_t$ contains a subgraph $H'$ which is a directed path of length $18m^2$, then construct a planning instance $\Pi_F$ according to the proof of Lemma 6, using the vertices from $H'$ as variables. Then, $CG(\Pi_F) \simeq H'$.

2. If $X_t$ contains a subgraph $H'$ which is a polypath with $m$ sinks and $m+1$ sources, then construct a planning instance $\Pi_F^-$ according to the proof of Lemma 7, using the variables of $H'$ as variables. Then, $CG(\Pi_F^-) \simeq F_m^{+1}$. This graph is a fence, i.e. a polypath where all directed paths are of length 1. Each such path can be 'stretched' to a directed path of arbitrary length by repeatedly applying Lemma 20. The graph $H'$ is a polypath that can be used as a template for which paths in $CG(\Pi_F^-)$ to stretch and how much in order to get a graph that is isomorphic to $H'$. Instance $\Pi_F^-$ can thus be modified into a new instance $\Pi_F$ such that $CG(\Pi_F) \simeq H'$.

All these constructions can be done in polynomial time, and for all cases, $\Pi_F$ has a solution if and only if $F$ is satisfiable. Furthermore, $CG(\Pi_F)$ is isomorphic to a subgraph of $H_t$ in all four cases. According to Lemma 17 it is thus possible to extend $\Pi_F$ to a new planning instance $\Pi_F^+$ such that $CG(\Pi_F^+) \simeq H_t$ and $\Pi_F^+$ has a solution if and only if $\Pi$ has a solution. This extension can be done in polynomial time according to the same lemma.

Since PlanExist($\mathcal{C}$) can be solved in polynomial time by assumption in the theorem, it thus follows that $M$ can solve 3SAT in polynomial time. However, this implies that $\mathbf{NP} \subseteq \mathbf{P}/\text{poly}$, which is impossible unless the polynomial hierarchy collapses (Theorem 16).

To see that the result holds even if the operators under consideration have at most 2 preconditions and 1 postcondition, simply note that this restriction holds for all reductions used in the underlying $\mathbf{NP}$-hardness proofs in Section 4. Similarly, the acyclicity restriction holds since the result is based only on in-stars, out-stars and polypaths, which are all acyclic graphs. □

Recall the generic blocks world encoding that we discussed in the beginning of this section. The class $\mathcal{D}$ of causal graphs for these blocks-world instances satisfies the requirements in Theorem 21, which means that PlanExist($\mathcal{D}$) is not likely to be tractable. However, finding non-optimal plans for blocks world is tractable; a plan of length at most twice the length of the optimal plan can be found in polynomial time (Gupta & Nau, 1992). That is, there are most likely more difficult problems than blocks world that happen to have exactly the same causal graphs, which illustrates that the complexity of a generic planning problem cannot be deduced from its corresponding class of causal graphs alone.

## 7. NP-Hard and NP-Intermediate Classes

The theorem by Chen and Giménez (2010) states a crisp complexity-theoretic borderline: if the component sizes are bounded by a constant, then planning is polynomial-time solvable and, otherwise, planning is not polynomial-time solvable. We have exploited an extra constraint, SP-closure, to be able to prove **NP**-hardness, which leaves a greyzone between the polynomial cases and the **NP**-hard ones. If we no longer require the classes to be SP-closed, then they are no longer obviously **NP**-hard even if the components are unbounded. The natural question then arises, can we say something about this middle ground? For instance, can we say something about what the **NP**-intermediate cases may look like and where the borderline between **NP**-hard and **NP**-intermediate is? Although it does not seem likely that we could find any results that characterize this borderline exactly, we can at least give some partial answers to these questions. We will do this by proving two theorems related to the growth rate of the components. The first of these shows that planning is still **NP**-hard if the components grow as $O(|V(G)|^{1/k})$ for integers $k$, while the second one shows that planning is likely to be **NP**-intermediate if the components grow polylogarithmically.

**Theorem 22.** *For every constant integer $k > 1$, there is a class $\mathcal{G}_k$ of graphs such that cc-size$(G) \leq |V(G)|^{1/k}$ for all $G \in \mathcal{G}_k$ and PlanExist$(\mathcal{G}_k)$ is **NP**-hard.*

*Proof.* Let $k > 1$ be an arbitrary integer. Construct the graph class $\mathcal{G}_k = \{G_1, G_2, G_3, \ldots\}$ as follows. For each $m > 0$, let $G_m$ have $m^{k-1}$ components, each of them isomorphic to $dP_m$, i.e. $|V(G_m)| = m^k$ so all components are of size $m = |V(G_m)|^{1/k}$. We prove **NP**-hardness of PlanExist$(\mathcal{G}_k)$ by reduction from PlanExist$(\mathbf{dP})$. Let $\Pi$ be an arbitrary planning instance such that $CG(\Pi) \in \mathbf{dP}$. Then $CG(\Pi) = dP_m$ for some $m > 0$. Construct a new instance $\Pi'$ which consists of $m^{k-1}$ renamed copies of $\Pi$. This is clearly a polynomial time construction since $k$ is constant and $m < ||\Pi||$. Furthermore, $CG(\Pi')$ is isomorphic to $G_m$ and $\Pi'$ has a solution if and only if $\Pi$ has a solution. Hence, this is a polynomial reduction so it follows from Lemma 6 that PlanExist$(\mathcal{G}_k)$ is **NP**-hard. □

Obviously, the size of the graphs is exponential in $k$.

Our second result must be conditioned by the assumption that the exponential time hypothesis (Impagliazzo & Paturi, 2001; Impagliazzo, Paturi, & Zane, 2001) holds. This hypothesis is a conjecture stated as follows.

**Definition 23.** *For all constant integers $k > 2$, let $s_k$ be the infimum of all real numbers $\delta$ such that $k$-SAT can be solved in $O(2^{\delta n})$ time, where $n$ is the number of variables of an instance. The exponential time hypothesis (ETH) is the conjecture that $s_k > 0$ for all $k > 2$.*

Informally, ETH says that satisfiability cannot be solved in subexponential time. ETH is not just an arbitrarily choosen concept, but a quite strong assumption that allows for defining a theory similar to the one of **NP**-completeness. There is a concept called SERF (subexponential reduction family) reduction which preserves subexponential time solvability. There is also a concept called SERF-completeness which is similar to **NP**-completeness, but based on SERF reductions. That is, there is a subclass of the **NP**-complete problems that are also SERF-complete, meaning that these can all be SERF reduced to each other. Hence, if one of these can be solved in subexponential time, then all of them can.

**Theorem 24.** *For all constant integers $k > 0$ and all classes $\mathcal{C}$ of directed graphs, if $\mathsf{cc\text{-}size}(G) \leq \log^k |V(G)|$ for all $G \in \mathcal{C}$, then $\mathsf{PlanExist}(\mathcal{C})$ is not $\mathbf{NP}$-hard unless ETH is false.*

*Proof.* Let $k > 0$ be an arbitrary integer. Let $\Pi$ be an arbitrary planning instance with $n$ variables of maximum domain size $d$ such that $\mathsf{cc\text{-}size}(CG(\Pi)) \leq c$. The components correspond to independent subinstances, which can thus be solved separately. Each component has a state space of size $d^c$ or less, so a plan for the corresponding subinstance can be found in $O(d^{2c})$ time, using Dijkstra's algorithm. Since there are at most $n$ components, the whole instance can be solved in $O(nd^{2c})$ time. However, it follows from the standard assumptions of reasonable encodings that both $n \leq ||\Pi||$ and $d \leq ||\Pi||$, so a looser bound is that $\Pi$ can be solved in $O(x \cdot x^{2c}) = O(x^{1+2c})$ time, where $x = ||\Pi||$.

Suppose $\mathsf{PlanExist}(\mathcal{C})$ is $\mathbf{NP}$-hard. Then there is a polynomial reduction from 3SAT to $\mathsf{PlanExist}(C)$. Furthermore, the size of a 3SAT instance is polynomially bounded in the number of variables. Hence, there must be some polynomial $p$ such that for a 3SAT instance with $n$ variables, the corresponding planning instance $\Pi$ has size $||\Pi|| \leq p(n)$.

Since the number of variables in $\Pi$ is upper bounded by $||\Pi||$, it follows from the assumption that the component size is upper bounded by $\log^k ||\Pi|| \leq \log^k p(n)$. Hence, $\Pi$ can be solved in $O(p(n)^{1+2\log^k p(n)})$ time, according to the earlier observation, and

$$p(n)^{1+2\log^k p(n)} = (2^{\log p(n)})^{1+2\log^k p(n)} \leq (2^{(1+2\log^k p(n))\log^k p(n)}) \leq 2^{3\log^{2k} p(n)}.$$

Furthermore, $\log^k p(n) \in O(\log^k n)$, since $p$ is a polynomial, so $2^{3\log^{2k} p(n)} \in 2^{O(\log^{2k} n)}$ and it follows that $\Pi$ can be solved in $2^{O(\log^{2k} n)}$ time. However, then $\Pi$ can be solved in $2^{\delta n}$ time for arbitrarily small $\delta$, which contradicts ETH. It follows that $\mathsf{PlanExist}(\mathcal{C})$ cannot be $\mathbf{NP}$-hard unless ETH is false. $\square$

Since the components are unbounded, this problem is not likely to be solvable in polynomial time either. It is thus an $\mathbf{NP}$-intermediate problem under the double assumption that $\mathbf{W}[1] \not\subseteq \mathbf{nu\text{-}FPT}$ and that ETH holds.

Theorems 22 and 24 together thus tell us something about where the borderline between $\mathbf{NP}$-intermediate and $\mathbf{NP}$-hard graph classes is. However, it is not a very crisp distinction; asymptotically, there is quite a gap between the polylogarithmic functions and the root functions (i.e. functions on the form $x^{1/k}$). One may, for instance, note that the function $f(n) = 2^{(\log n)^{1-\frac{1}{(\log \log n)^c}}}$ lies within this gap whenever $0 < c < 1$.

## 8. Discussion

SP-closed graph classes have appealing properties and fit in well as a concept stronger than subgraph-closed but weaker than minor-closed. They also give a partial characterization of where the borderline to $\mathbf{NP}$-hardness lies. However, as noted earlier, it is possible to define other types of graph classes which also imply that planning is $\mathbf{NP}$-hard. One example is the family $\mathcal{G}_1, \mathcal{G}_2, \mathcal{G}_3, \ldots$ of classes in the proof of Theorem 22. Another more specialized and, perhaps, contrived class is the following, intended to give a contrast to the SP-closure concept and the $\mathcal{G}_k$ classes.

A *tournament* is a directed graph formed by giving directions to each edge in a complete graph. Let **T** denote the set of tournaments and note that **T** is *not* SP-closed. However, tournaments are Hamiltonian graphs (Rédei, 1934) so if $T$ is a tournament on $n$ vertices, then path-length$(T) = n - 1$. Furthermore, the path of length $n - 1$ can be computed in polynomial time (Bar-Noy & Naor, 1990).

Assume we are given a 3SAT formula $F$ with $n$ variables and $m$ clauses. Let $\ell = (2m + 4)n$, i.e. $\ell$ is polynomially bounded in $F$. According to Lemma 6 we can thus construct a planning instance $\Pi_F$ in polynomial time such that

1. $\Pi_F$ contains $\ell$ variables,

2. $CG(\Pi_F) \simeq dP_\ell$, and

3. $\Pi_F$ has a solution if and only if $F$ is satisfiable.

Choose an arbitrary tournament $T$ with $\ell$ vertices in **T**. Find the path of length $\ell - 1$ in $T$ and identify it with $CG(\Pi_F)$. Then add dummy operators corresponding to the remaining edges of $T$. We have thus shown that there is a polynomial-time transformation from 3SAT to PlanExist(**T**), and that PlanExist(**T**) is **NP**-hard. One may also note that variations of this technique can be used for proving that PlanExist(**T**′) is **NP**-hard for many different **T**′ ⊆ **T**.

While we have not considered domain sizes or tractable restrictions in this article, we note that the Theorem 24 may give some ideas for where to look for tractable cases. Consider the case where all variable domains are bounded in size by some constant $k$ and where cc-size$(G) \leq \log V(G)$. Using the first part of the proof, we see that planning can be solved in $O(n \cdot k^{2\log n})$ time. However, $k^{2\log n} = (2^{\log k})^{2\log n} = (2^{\log n})^{2\log k} = n^{2\log k}$, which is polynomial since $k$ is a constant. That is, planning is tractable for this restricted case. Even though this observation is straightforward, it is interesting as a contrast to Theorem 24. It also suggests that there are even larger tractable subgraphs if we also consider additional restrictions on the planning instances.

While we have explicitly commented on the sufficient number of pre- and postconditions for the various results, there are also alternative such characterizations that might be relevant. It would bear to far to list all such possibilities, so let it suffice with one example. The concept of *prevail conditions*, i.e. preconditions on variables that are not changed by the operator, originate from the SAS$^+$ formalism (Bäckström & Nebel, 1995) but has more recently been considered also in the context of causal graphs. Giménez and Jonsson (2012) refer to an operator as *k-dependent* if it has a precondition on at most $k$ variables that it does not also change. We may note that the proofs of Lemmata 17 and 20 only introduce operators that are 1-dependent, at most. Since the proof of Theorem 21 does not impose any further such restrictions on the original planning instance, it follows that this theorem holds also when all operators are 1-dependent, at most.

As a final question, one might wonder if it is of any practical use at all to know that planning is tractable, or **NP**-intermediate, for severely limited component sizes? After all, most planning instances are likely to have a causal graph that is weakly connected, that is, the whole graph is one single component. To answer that question, the first important observation to make is that the complexity of planning for instances is directly related to the complexity of planning for the components separately. This is because there can be at most linearly (in the number of variables) many components. If planning can be solved in polynomial time for all components of an instance, then it can be solved in polynomial time for the whole instance. Conversely, if planning cannot be solved in polynomial time for the whole instance, then there is at least one component which is not polynomial-time solvable. That is, the complexity results for instances and for components are directly related to each other. In other words, the results are relevant for all methods that artificially split the causal graph into components, in one way or another. Examples are the causal-graph heuristic by Helmert (2006a), factored planning (Brafman & Domshlak, 2006) and structural pattern data bases (Katz & Domshlak, 2010).

## Acknowledgments

The anonymous reviewers provided valuable comments and suggestions for improving this article.